\documentclass[acmtog]{acmart}
\acmSubmissionID{2542} 

\AtBeginDocument{%
  \providecommand\BibTeX{{%
    \normalfont B\kern-0.5em{\scshape i\kern-0.25em b}\kern-0.8em\TeX}}}

\usepackage{amsmath,amsfonts,bm}
\usepackage{multirow}
\usepackage{graphicx}
\usepackage{xspace}
\usepackage{titletoc}   

\usepackage{algorithm}

\newcommand{\mset}[1]{\left\{\kern-.5em\left\{ #1 \right\}\kern-.5em\right\}}
\newcommand{\mmset}[1]{\{\kern-.4em\{ #1 \}\kern-.4em\}}

\newcommand{\myparagraph}[1]{\vspace{0.15cm}\noindent{ \bf{\emph{ #1}\hspace{0.05cm}}}}

\def\eqref#1{Eq.~\ref{#1}}

\def\1{\bm{1}}

\def\vec1{{\bm{1}}}

\DeclareMathAlphabet{\mathsfit}{\encodingdefault}{\sfdefault}{m}{sl}
\SetMathAlphabet{\mathsfit}{bold}{\encodingdefault}{\sfdefault}{bx}{n}

\newcommand{\mytextapprox}{\raisebox{0.5ex}{\texttildelow}}
\usepackage{makecell}
\usepackage{booktabs}
\usepackage{enumitem}
\usepackage{tabularx}
\usepackage{algpseudocode}
\usepackage{cleveref}
\usepackage{soul}

\copyrightyear{2025}
\acmYear{2025}
\setcctype{by}
\acmConference[SA Conference Papers '25]{SIGGRAPH Asia 2025 Conference Papers}{December 15--18, 2025}{Hong Kong, Hong Kong}
\acmBooktitle{SIGGRAPH Asia 2025 Conference Papers (SA Conference Papers '25), December 15--18, 2025, Hong Kong, Hong Kong}\acmDOI{10.1145/3757377.3764008}
\acmISBN{979-8-4007-2137-3/2025/12}

\begin{document}
\title{DynVFX: Augmenting Real Videos with Dynamic Content}

\author{Danah Yatim}
\authornote{Both authors contributed equally to this research.}
\affiliation{%
 \institution{Weizmann Institute of Science}
 \country{Israel}
 }
\author{Rafail Fridman}
\authornotemark[1]
\affiliation{%
 \institution{Weizmann Institute of Science}
 \country{Israel}
 }
\author{Omer Bar-Tal}
\affiliation{%
 \institution{Runway ML}
 \country{United States of America}
 }
\author{Tali Dekel}
\affiliation{%
 \institution{Weizmann Institute of Science}
 \country{Israel}
 }
\newcommand{\bs}[1]{\boldsymbol{#1}}
\newcommand{\afterfigure}{\vspace{-3mm}} 
\begin{abstract}
We present a method for augmenting real-world videos with newly generated dynamic content. Given an input video and a simple user-provided text instruction describing the desired content, our method synthesizes dynamic objects or complex scene effects that naturally interact with the existing scene over time. The position, appearance, and motion of the new content are seamlessly integrated into the original footage while accounting for camera motion, occlusions, and interactions with other dynamic objects in the scene, resulting in a cohesive and realistic output video. We achieve this via a zero-shot, training-free framework that harnesses a pre-trained text-to-video diffusion transformer to synthesize the new content and a pre-trained vision-language model to envision the augmented scene in detail. Specifically, we introduce a novel inference-based method that manipulates features within the attention mechanism, enabling accurate localization and seamless integration of the new content while preserving the integrity of the original scene.
Our method is fully automated, requiring only a simple user instruction. We demonstrate its effectiveness on a wide range of edits applied to real-world videos, encompassing diverse objects and scenarios involving both camera and object motion\footnote{Code will be made publicly available.}.
Project page: \url{https://dynvfx.github.io/}
\end{abstract}



\begin{teaserfigure}
  \includegraphics[width=\textwidth]{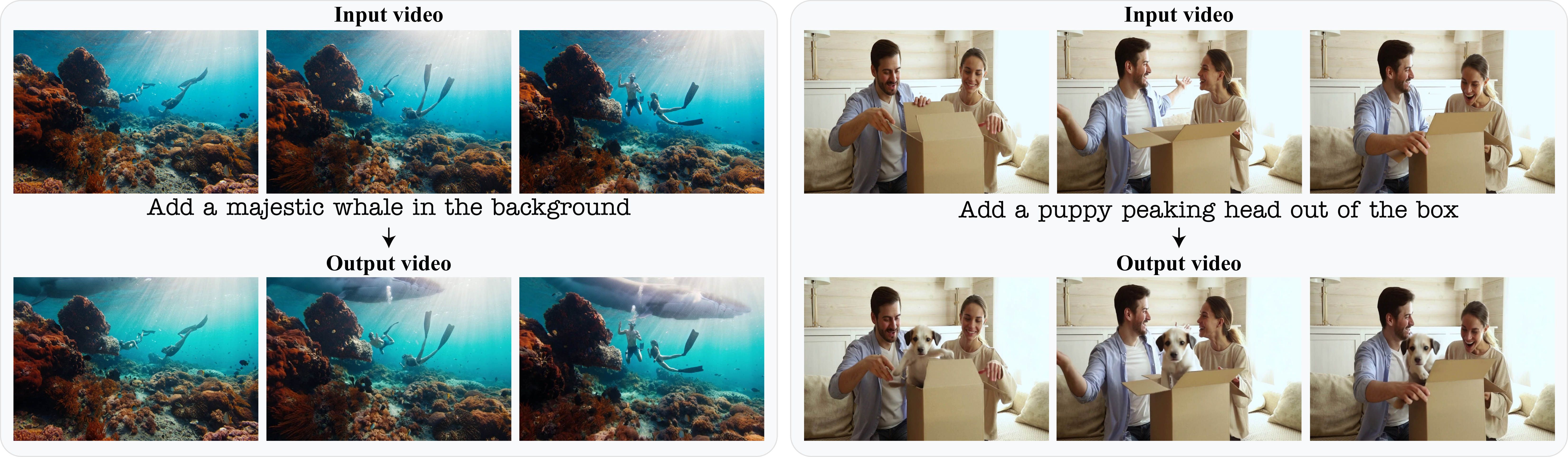} \vspace{-0.5cm}
  \caption{DynVFX augments real-world videos with new dynamic content described by a simple user-provided text instruction (shown in the top row under each input video). The framework automatically infers where the synthesized content should appear, how it should move, and how it should harmonize at the pixel level with the scene, without requiring any additional user input. The key idea is to selectively extend the attention mechanism in a pre-trained text-to-video diffusion model, enforcing the generation to be content-aware of existing scene elements (anchors) from the original video. This allows the model to generate content that naturally interacts with the environment, producing complex and realistic video edits in a fully automated way (bottom row).}
  \label{fig:teaser}
\end{teaserfigure}

\maketitle

\section{Introduction}
\label{sec:intro}

Incorporating computer-generated imagery (CGI) into real-world footage has been a transformative capability in film production,  enabling the creation of visual effects that would be difficult or impossible to achieve otherwise. 
For instance, the seamless integration of CGI characters, such as \emph{Gollum} in \emph{The Lord of the Rings} or  \emph{T-Rex} in \emph{Jurassic Park}, has empowered filmmakers to blend fantastical elements with real-world environments, resulting in immersive storytelling. Professional video editing software tools provide artists with precise control over visual effects integration into video footage \cite{maya,blender,unrealengine,adobe}. However, they require significant expertise, manual effort, pre-made assets, and financial resources. To disentangle these inspiring storytelling capabilities from professional and economic status, we propose a new creative task: free-vocabulary, text-driven integration of newly generated dynamic content into real-world videos. Specifically, given an input video along with a short user instruction describing the new content (e.g., ``add a massive whale''), our goal is to synthesize new dynamic objects or complex scene effects that naturally interact with the existing scene across the entire video (Fig.~\ref{fig:teaser}). Importantly, our focus is to achieve this without additional user input, allowing anyone to insert dynamic content into videos just by describing it.

We approach this fully automated task by inferring where the synthesized content should appear, how it should move, and how it should harmonize at the pixel level with the scene without requiring any additional user input. This is achieved by harnessing dynamic generative priors learned by a pre-trained text-to-video diffusion transformer (DiT) model and comprehensive scene understanding learned by a vision-language model without any fine-tuning or additional training. This contrasts with previous methods that rely on additional user input, such as VFX assets \cite{hsu2024autovfx} or per-frame references to specify the placement of the new content over time. These methods struggle with integrating complex dynamic content (e.g., synthesizing multiple articulated objects or global effects, as shown in the dinosaurs and tsunami examples in Fig. \ref{fig:results}).

Our task poses several new fundamental challenges. First, the generation process of the fixed pre-trained DiT model must be \emph{content-aware} such that the position, appearance, and motion of the synthesized dynamic content integrates naturally with the original scene. This entails synthesizing objects that respect occlusions, maintain appropriate relative size and perspective relative to the camera position, and realistically interact with other dynamic objects. All of this must be achieved while maintaining the integrity of the original video, ensuring that new content enhances the scene without compromising its authenticity. 

Existing text-based methods focus on controlling diffusion generation to preserve original scene aspects. However, this imposes a trade-off between preserving aspects
of the original scene and adding new content to the scene: either over-fitting or under-fitting the original video in terms of appearance, motion, positioning, or scale. Attention manipulation within a UNet-based architecture has been widely used for content preservation in video editing tasks. Extending attention across frames in these models allows for global appearance coherence. Yet, spatiotemporal information is preserved only implicitly. Hence, in the context of UNet-based models, this has been used mainly for appearance transfer. We revisit extended attention in the context of DiT-based text-to-video models, revealing new insights into their internal representation. Unlike UNet-based models, we observe that keys/values in DiTs locally encode corresponding video patches across space and time. In particular, applying the same positional embedding to both source and target keys/values during the extended attention operation enables alignment between the generated and original content at corresponding spatiotemporal locations. 




We take advantage of this property to enforce content-aware generation for new content placement.
Specifically, we steer the localization of the edit through \textit{Anchor Extended Attention}: we incorporate a specific set of keys/values extracted from the original video as additional context to the DiT-based model during sampling, enabling the model to focus on existing scene elements essential for correctly placing and integrating new content naturally.

Despite accurate placement from \textit{Anchor Extended Attention}, it does not guarantee pixel-level alignment with the original scene. To achieve better harmonization, we iteratively refine the edit by repeatedly sampling with \textit{Anchor Extended Attention} while updating only the edited regions in each iteration.

To allow a fully automated framework, a vision-language model is utilized to interpret the interaction and reason about integration. We guide the model to: (1) translate the user's instructions into detailed scene descriptions and (2) identify both the prominent visual elements of the existing scene and the new content to be introduced. These identified objects are used in our framework for the application of a text-based segmentation model to aid the localization and harmonization of the new content.

We demonstrate the effectiveness of our approach on a variety of edits applied to real-world videos. Our method supports a wide range of scene augmentations across various scenarios while maintaining realistic interaction, occlusion, lighting, and camera motion.
It is worth noting that our method utilizes a publicly available text-to-video model, which exhibits a significant gap in video generation quality compared to recent state-of-the-art video models. Nevertheless, we observe that within our problem formulation and objectives, we can distill surprisingly powerful generative capabilities from this model.

To summarize, our work makes the following contributions:
\begin{itemize}[topsep=1pt,partopsep=0pt,leftmargin=.5cm]
    \item We introduce a new task of integrating newly generated dynamic content into real-world videos without relying on the user to provide complex references for the effect.
    \item We propose a tuning-free, zero-shot, fully automated method that enables harmonized content integration while maintaining high fidelity to the original scene. 
    \item We demonstrate state-of-the-art results, achieving the best trade-off between synthesizing new dynamic elements and maintaining high fidelity to the original content.
\end{itemize}

\section{Related Work}


\myparagraph{Leveraging Text-to-Image Models.}
With the advancement of text-to-image (T2I) models, techniques for image manipulation based on such models have evolved rapidly. Among these advancements, notable progress has been made in the task of instruction-based image editing. Several works \cite{hive,Sheynin2023EmuEP,brooks2022instructpix2pix,Zhang2023MagicBrush} have proposed directly fine-tuning generative models on pairs of original and edited images coupled with user-provided instructions. 

Object insertion in images falls under the broader umbrella of instruction-based editing techniques. Recent object insertion methods \cite{erasedraw-24, Wasserman2024PaintBI, winter2024objectdrop} leverage inpainting models to create paired image datasets, which are then used to fine-tune image editing models. However, extending these approaches to videos presents significant challenges. In particular, generating large-scale instruction-paired video datasets can be expensive in both time and computational resources, as it requires manual effort to annotate frames. This cost and complexity make it challenging to adapt existing image-based methodologies directly to the video domain.

Several works explore the manipulation of internal feature representation in T2I models for image and video editing tasks.
In U-Net–based architectures, attention keys and values were shown to encode global appearance information \cite{masactrl}, which several video editing methods \cite{qi2023fatezero,wu2022tuneavideo,text2video-zero} exploit by extending self-attention across frames to enforce global appearance consistency. For localized appearance coherency, MasaCtrl \cite{masactrl} and Consistory \cite{tewel2024consistory} mask and selectively replace or extend attention keys and values. Add-It \cite{Tewel2024AdditTO} employs global weighted extended attention for object insertion in images. FlowEdit \cite{kulikov2024flowedit} uses pre-trained flow models for inversion-free local edits.
In contrast to the methods mentioned above that may fail to preserve the original content, we propose applying extended attention only to specific regions of the source scene, allowing generation to focus on essential elements. 

\begin{figure*}[ht!]
    \centering
    \includegraphics[width=\textwidth]{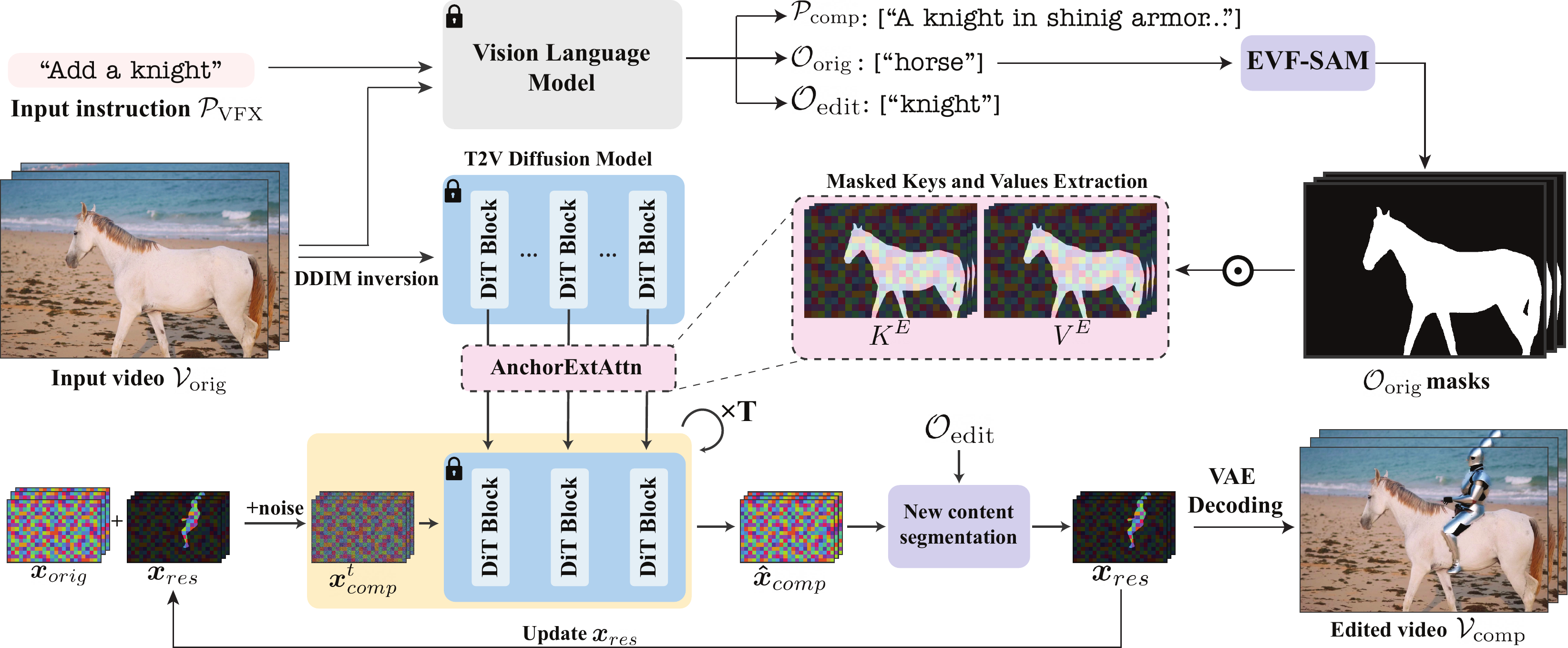}
    \caption{{\bf Pipeline.} \emph{Top (pre-processing).} Given an input video $\mathcal{V}_{\text{orig}}$ and instruction $\mathcal{P}_{\text{VFX}}$, we \textit{(i)} Apply our VLM protocol for instruction interpretation, to yield a comprehensive scene description $\mathcal{P}_{\text{comp}}$, original objects $\mathcal{O}_{\text{orig}}$ and target object $\mathcal{O}_{\text{edit}}$ descriptions. \textit{(ii)} DDIM invert $\mathcal{V}_{\text{orig}}$ to extract spatiotemporal keys/values $[\mathbf{K}_\text{orig}, \mathbf{V}_\text{orig}]$. \emph{Bottom (editing).} We initialize the composed latent with $\bs{x}_{\text{comp}}=\bs{x}_{\text{orig}}$ and iterate over a list of descending noise levels $t=\tilde{T}\!\rightarrow\!T_{\min}$ used for noising $\bs{x}_{\text{comp}}$. At each iteration $t$ we: \textit{(i)} noise $\bs{x}_{\text{comp}}$ to noise level $t$, and sample with Anchored Extended Attention, to output $\bs{\hat{x}_{\text{comp}}}$. \textit{(ii)} Update $\bs{x}_{\text{comp}}$ within the new contents masked regions $\bs{{M}_{\text{VFX}}}$ by 
adding the residual $\bs{x}_{\text{res}}=\bs{{M}_{\text{VFX}}\cdot(\hat{x}_{\text{comp}}-x_{\text{orig}}})$ to $\bs{x}_{\text{orig}}$. Repeating this loop gradually integrates the new content, yielding the edited video $\mathcal{V}_{\text{comp}}$.
     }
    \label{fig:pipeline}\afterfigure
\end{figure*}
 
\myparagraph{Controllable Video Generation and Editing.}
Recently, numerous methods have been developed to incorporate various forms of control signals into video generation pipelines.
Several video-to-video methods propose to condition the generation on per-frame spatial maps such as depth maps and edge maps \cite{Wang2023VideoComposerCV,Chen2023ControlAVideoCT}. A common approach in video editing is to enforce the preservation of the original scene layout and motion for various tasks. Several consistent video editing works \cite{bar2022text2live, lee2023textvideoedit, tokenflow2023, cong2023flatten} restrict the preservation of scene structure and motion, and rely on a 2D generative prior. Therefore, they cannot support large structural deviations or motion adaptation.  \cite{yatim2023spacetime, zhao2023motiondirector, park2024spectral, jeong2023vmc, materzynska2024newmove} proposed to utilize a text-to-video (T2V) model for the task of motion transfer by controlling diffusion video generation to preserve original scene motion. 

\cite{ku2024anyv2v, ouyang2024i2vedit} adopt a two-stage pipeline: first, apply an off-the-shelf image editing method to the video’s first frame, then use an image-to-video generation model to propagate that edit through time via temporal feature injection.
All of these methods fail to insert new dynamic content. Our approach focuses on integrating additional dynamic elements into the video. 

\myparagraph{Reference-Based Video Content Insertion.}
Recent methods \cite{bartal2024lumiere,Ma2024VidPanosGP,Zhang2023AVIDAV,Zi2024CoCoCoIT, zhou2023propainter, mou2024revideoremakevideomotion, tu2025videoanydoorhighfidelityvideoobject, bian2025videopainter} have explored the adaptation of video models for video inpainting by conditioning on a masked video and a corresponding binary mask. This setup encourages the model to preserve unmasked information while generating new content in the masked region. Some methods, such as ReVideo \cite{mou2024revideoremakevideomotion} and VideoAnydoor \cite{tu2025videoanydoorhighfidelityvideoobject}, additionally guide the motion of the new content based on user-provided motion trajectories and bounding boxes.

VideoDoodles \cite{videodoodles} combines hand-drawn animations with video footage in a scene-aware manner by tracking a user-provided planar canvas in 3D. However, it requires key-frame edit placement. Our task fundamentally differs from these methods, which all require explicit per-frame control over where (via per-frame masks \cite{bian2025videopainter} or bounding boxes \cite{mou2024revideoremakevideomotion}) or how the new content will move (via motion trajectories \cite{mou2024revideoremakevideomotion}). In contrast, our method does not constrain either aspect and requires only a simple text instruction as input. The locations and dynamics in our framework emerge from the generative prior of the T2V model.
Furthermore, the reliance of these methods on user-provided per-frame annotations is impractical for integrating complex dynamics (e.g., Fig. \ref{fig:results} tsunami and dinosaurs). While a static object can be masked with a simple bounding box, manually defining per-frame masks for complex motion or interactions is extremely difficult. \\
Other approaches \cite{kling, pika} reference the asset itself by requiring the user to additionally input an image of the object. This approach limits their applicability whenever a suitable example cannot be found for the given video. For instance, these methods struggle to generate global effects, making them less preferable than text-based insertion methods.

\myparagraph{Language Models for Video Content Creation.}
Advancements in Vision-Language Models (VLMs) have enabled
methods to utilize such models in various video-related tasks. Some methods \cite{Chen2024Panda70MC7,yang2024cogvideox} use VLMs to produce detailed video captions from a series of frames, which are then utilized to train T2V generative models.
Other methods utilize such models for achieving better generation controllability. For instance, VideoDirectorGPT \cite{Lin2023VideoDirectorGPT} utilizes a VLM for multi-scene video generation by training diffusion adapters to incorporate additional conditioning inputs, while LVD \cite{lian2023llmgroundedvideo} incorporates layout guidance from the VLM during the sampling process. AutoVFX \cite{hsu2024autovfx} uses an LLM to generate a video editing program pipeline based on the user's instructions. In our work, we employ a VLM as a ``VFX assistant''. Based on a short user instruction, it yields a comprehensive description of the edited video along with the prominent objects present in the scene.


\section{Preliminaries}
\label{sec:preliminaries}
Diffusion probabilistic models \cite{ddpm,sohl2015deep} are a class of generative models that aim to learn a mapping from noise $\bs{x}_T\sim \mathcal{N}(0,I)$ to a data distribution $q$. Starting from a Gaussian i.i.d. noise sample $\bs{x}_T\sim \mathcal{N}(0,I)$, the diffusion model $\Phi$ is applied iteratively through a sequence of denoising steps, ultimately producing a clean output sample $\bs{x}_0$. 

Recently, a new class of latent T2V models, built on Diffusion Transformers (DiTs) \cite{dit}, has gained significant popularity, as DiTs enhance spatial coherence and enable training for arbitrary aspect ratios and video lengths. Unlike previous U-Net–based architectures, which utilize interleaving 2D spatial attention and 1D temporal attention layers, DiT treats the entire video as a long token sequence, where both text and spatiotemporal tokens are processed with full self-attention layers, a mechanism often referred to as 3D attention. In each DiT block, text tokens and spatiotemporal tokens are projected into queries, keys, and values using separate sets of weights for each modality, and the sequences of the two modalities are concatenated as a joint input for the attention operation. The attention \cite{Vaswani2017AttentionIA} operation computes the affinities between the d-dimensional projections $\mathbf{Q}$ and $\mathbf{K}$ to yield the output of the layer: 
\begin{equation}
\mathbf{A}\cdot\mathbf{V} \ \text{where} \
\mathbf{A} = \texttt{Attention}(\mathbf{Q}, \mathbf{K}) = \texttt{softmax}\left( \frac{\mathbf{Q} \mathbf{K}^\top}{\sqrt{d}}\right)\, .
\label{eq:attnetion}
\end{equation}
To capture inter-token relationships, Rotary Position Embeddings (RoPE) \cite{Su2021RoFormerETrope} are applied to the input queries and keys in the attention operation.
In this work, we utilize a publicly available DiT-based model \cite{yang2024cogvideox}, CogVideoX, for augmenting real-world videos with newly generated dynamic content in a zero-shot manner. 


\begin{figure*}[h]
    \centering
    \includegraphics[width=\textwidth]{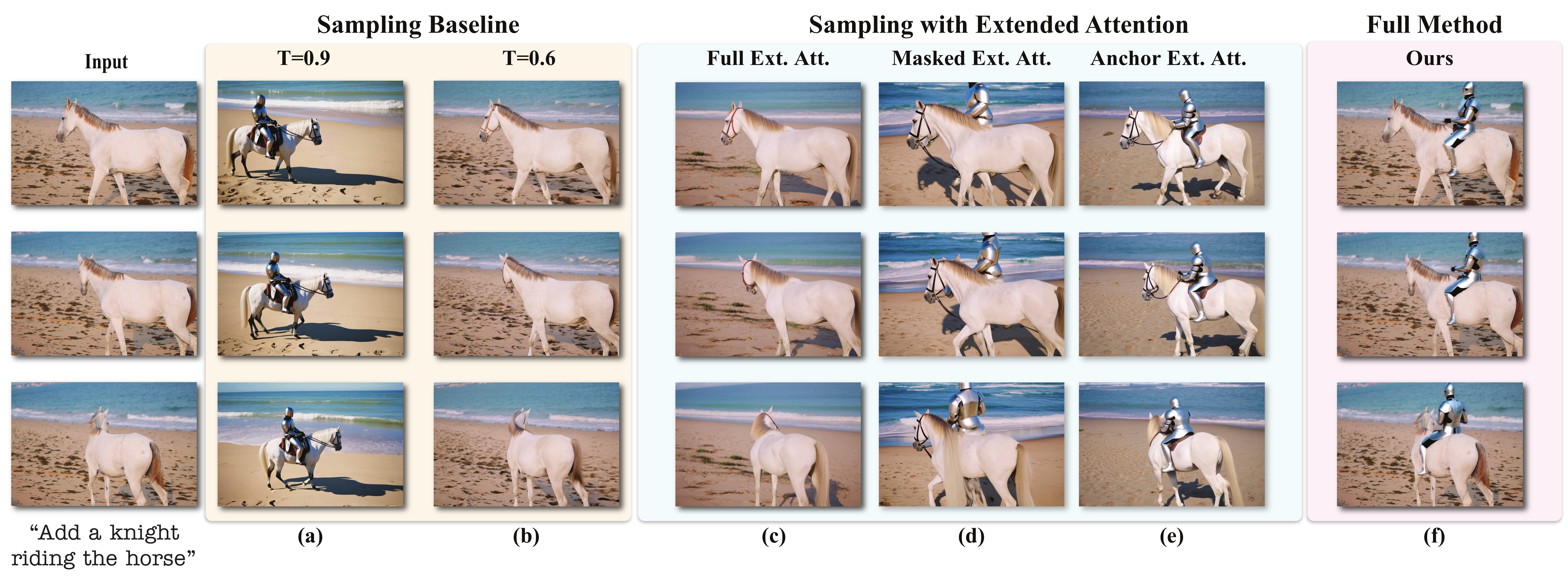} \vspace{-0.8cm}
    \caption{{\bf Controlling Fidelity to the Original Scene Using Different Extended Attention Mechanisms.} (a-b) SDEdit suffers from the original scene preservation/edit fidelity trade-off. (c-e) Three Extended Attention variants during sampling demonstrate different control levels: Full Extended Attention closely reconstructs the input scene, Masked Extended Attention proves too constrained in overlapping regions despite allowing new content emergence, and our Anchor Extended Attention. achieves optimal results by applying dropout -- extending attention only at sparse points within selected regions.}
    \label{fig:extended_attn}\afterfigure
\end{figure*}

\section{Method}
\label{sec:method}
Given an input video $\mathcal{V}_{\text{orig}}$ and a textual instruction $\mathcal{P}_{\text{VFX}}$, our goal is to synthesize a new video $\mathcal{V}_{\text{VFX}}$ in which new dynamic elements are seamlessly integrated into the existing scene.
Tackling this task requires ensuring that the location and size of the new content align with the camera motion and the environment, while its actions and movements must respond appropriately to other dynamic objects present in the scene. Our framework (Fig.~\ref{fig:pipeline}) addresses these challenges by incorporating the following key components:

\begin{enumerate}[leftmargin=.5cm]
    \item \textit{VLM as a VFX Assistant.} We utilize a pre-trained VLM to interpret the user’s instructions, reason about the interactions with the scene's dynamics, and identify both the prominent existing elements in the scene and the new content to be added. The VLM provides a descriptive prompt of the desired scene augmentation, a list of prominent existing elements, and the new content to be added. We achieve this by guiding the VLM to act as a ``VFX assistant'' via a system prompt containing guidelines in the context of our tasks.

    \item \textit{Localization via Anchor Extended Attention.} Our approach achieves precise placement of new content by guiding the T2V generation model during sampling to be content-aware. We steer the localization of the edit by applying Extended Attention with a sparse set of spatiotemporal locations extracted from the original scene, allowing the generation to focus on essential elements. Our approach is based on our observation that in the pre-trained T2V DiT model, keys and values locally encode corresponding video patches in both space and time.
    
    \item \textit{Content Harmonization via Iterative Refinement.} To guarantee precise pixel-level alignment of the generated content with the input video and achieve better harmonization, we iteratively refine the estimated edit by repeating the sampling process with AnchorExtAttn multiple times, while preserving all unedited regions.
\end{enumerate}

Our method does not require user-provided masks to pre-determine the location of the new content, but rather infers it automatically. We leverage a text-based segmentation model \cite{zhang2024evfsamearlyvisionlanguagefusion} for two purposes: \textit{(i)} Localization, by segmenting prominent elements from the original scene to apply AnchorExtAttn, which allows the T2V model to infer the natural location and dynamics of the new content based on the original scene context. \textit{(ii)} Harmonization, by segmenting the newly added content to ensure pixel-level alignment with the original scene.

\subsection{VLM as a VFX Assistant}
\label{sec:vlm}
To create a fully automatic framework that requires only a simple user instruction, we incorporate a VLM into our framework. Specifically, given a user instruction $\mathcal{P}_{\text{VFX}}$, along with keyframes from $\mathcal{V}_{\text{orig}}$, we instruct the VLM \cite{achiam2023gpt4} to provide a composition prompt $\mathcal{P}_{\text{comp}}$ describing the newly augmented scene. We use $\mathcal{P}_{\text{VFX}}$ as the edit prompt for the T2V generation process. Hence, the edit prompt must include the following details: (i) original scene elements and actions, (ii) new content elements and actions, and (iii) the interaction between them. When simply asking the VLM to yield $\mathcal{P}_{\text{VFX}}$, it often fails to produce captions that are both true to the original video and descriptive of the new content well. Therefore, using these for T2V generation can result in edits with low fidelity to the instruction (see examples in Sec. \ref{app:ablate_vfx_protocol} and Fig. \ref{fig:ablate_protocol} in the SM).


To resolve this issue, we guide the VLM in an in-context manner by instructing it via a system prompt to imagine a conversation with a VFX artist to obtain a caption that would describe the composited scene correctly. We observe that this helps the VLM to achieve better scene understanding. Specifically, the system prompt includes guidelines to focus on: (1) spatial and dynamic awareness of existing scene elements, (2) preservation of original scene behaviors, and (3) atmospheric coherence between new and existing content.

Additionally, we utilize the VLM to yield a list of prominent foreground objects in the original video $\mathcal{O}_{\text{orig}}$ and the object that will be added according to the edit prompt $\mathcal{O}_{\text{edit}}$. See additional details in Sec. \ref{app:obtain_morig}, Sec. \ref{app:latent_mask_extract}, and Sec. E in the SM. In addition, Figs. \ref{fig:protocol}-\ref{fig:vlm_evaluation} in the SM include an overview of the protocol, system prompts, and output examples.

\subsection{Localization via Anchor Extended Attention} 
\label{sec:method_localization}

A pivotal aspect of our method is the accurate placement of the new content. While $\mathcal{P}_{\text{comp}}$ can describe the desired location, naive noising-denoising with this prompt introduces a trade-off: As shown in Figs. \ref{fig:extended_attn} (a)-\ref{fig:extended_attn}(b), using SDEdit \cite{meng2022sdedit} with a high noising timestep fails to retain the original scene, resulting in misaligned new content, whereas a low noising timestep limits deviations from the original video.

To tackle the localization challenge, we extend the attention module during sampling to include the input video's corresponding attention features. Specifically, we apply DDIM inversion \cite{song2020_ddim} to the original video $\mathcal{V}_{\text{orig}}$ and extract the spatiotemporal keys and values $K_\text{orig}, V_\text{orig}$ from the attention module of every block in the network and generation timestep $t$. These keys and values are then used to extend the attention mechanism during sampling with $\mathcal{P}_{comp}$ to control the localization of the edit.

When using all keys and values, the extended attention operation can be expressed as:
\begin{equation}
\texttt{Attn}(Q_\text{VFX},[K_\text{VFX}, K_\text{orig}], [V_\text{VFX}, V_\text{orig}])\, .
\label{eq:ExtAttn}
\end{equation}

We observe that extending attention to the full set of keys and values approximately reconstructs the original video (Fig. \ref{fig:extended_attn}(c)). 
This demonstrates that the keys and values of attention layers in the DiT-based model determine more than just appearance information. In contrast to UNet-based architectures, keys and values locally encode corresponding video patches. We hypothesize that this occurs because the same positional embedding is applied to $K_\text{orig}$ as to $K_\text{VFX}$. Remarkably, applying the same positional embedding to both source and target keys and values enables alignment between the generated and original content at corresponding spatiotemporal locations.


\begin{figure*}[ht!]
    \centering
\includegraphics[width=0.93\textwidth]{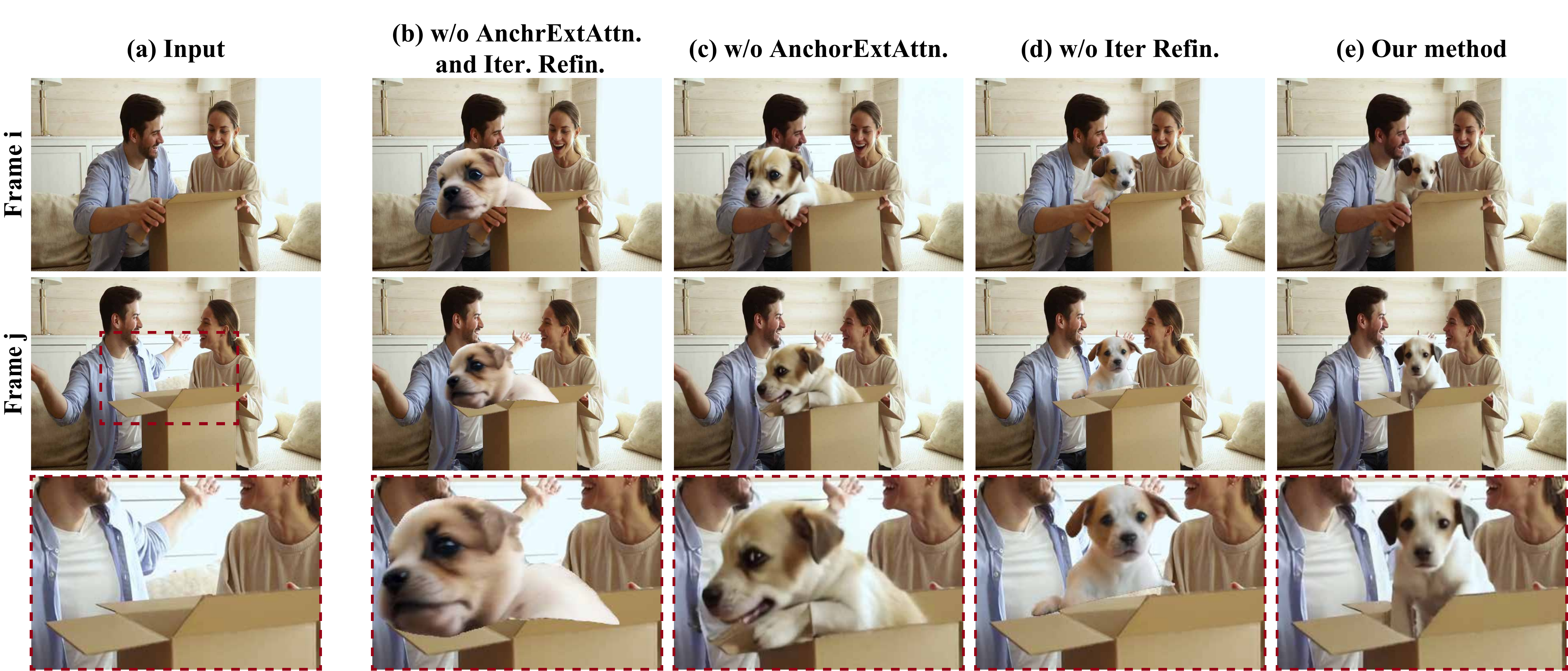}
\vspace{-0.1cm}
    \caption{Ablations. (b) Excluding both AnchorExtAttn and the Iterative refinement process results in significant misalignment with the original scene and poor harmonization (e.g., the size of the puppy relative to the scene and boundary artifacts). (c) Omitting AnchorExtAttn leads to incorrect positioning of the new content. (d) Removing iterative refinement results in poor harmonization. Our full method (e) exhibits good localization and harmonization of the edit.}
    \label{fig:ablation}\afterfigure
\end{figure*}

\myparagraph{Selective Attention.} 
Based on our observation, we propose restricting the extended attention only to specific positions in the source scene. Specifically, we use a selection function $\mathcal{F}$ to determine which keys and values are retained.

\textit{Masked Extended Attention.} A straightforward choice is region‐based masking: identifying the most critical regions for preserving scene coherence and extending attention with keys and values within these masked regions $M_{\text{orig}}$, i.e., $\mathcal{F}(x) = M_\text{orig} \circ x $.

To get $M_{\text{orig}}$, we ask a VLM for $\mathcal{O}_{\text{orig}}$, a list of foreground elements in $\mathcal{V}_{\text{orig}}$ (typically the most spatially prominent elements within a scene), and obtain the corresponding masks using a text-based segmentation model (see Sec. \ref{app:obtain_morig} and Sec. \ref{app:latent_mask_extract} in the SM for more details and examples of $M_{\text{orig}}$).

As shown in Fig.~\ref{fig:extended_attn}(d), this strategy successfully preserves high fidelity to the original scene within $M_{\text{orig}}$ while allowing new content to emerge in unmasked regions. However, Fig.~\ref{fig:extended_attn}(d) reveals two issues: (i) due to the hard constraint in overlapping regions, it is not suitable for handling natural occlusions, and (ii) the lack of guidance in unmasked regions can lead to inaccurate scale or positioning relative to the input scene.

\textit{Anchor Extended Attention.} To achieve robust and spatially coherent integration of new content, we propose to selectively extend keys/values in a sparse set of foreground anchors and a sparser set of background anchors. Introducing Anchor Extended Attention with its formulation:
{\small \begin{equation}
\begin{aligned}
  \texttt{AnchorExtAtt} := &\texttt{Attn}(Q_{\text{VFX}},[K_{\text{VFX}}, K^E],
  [V_{\text{VFX}}, V^E])\, , \\
  &\text{s.t. }K^E:=\mathcal{F}(K_{\text{orig}}) ~~and~~  V^E:= \mathcal{F}(V_{\text{orig}})\, , \\
  & \mathcal{F}(x) := \texttt{Drop}_{\text{FG}}(M_{\text{orig}}) \cup\texttt{Drop}_{\text{BG}}(\sim M_{\text{orig}}) ~\circ~ x\, .
\end{aligned}
\label{eq:AnchorExtAttn}
\end{equation}}
For each layer, a selection of anchor points is randomly sampled via dropout, in both masked and unmasked regions. 
To steer the localization of the edit, we extend the attention with anchors that correspond to spatiotemporal locations in $\mathcal{V}_{\text{orig}}$, allowing the generation to focus on essential elements (anchors) to achieve robust and spatially coherent integration of new content. (e.g., the knight's motion is aligned with the horse in Fig.~\ref{fig:extended_attn}(e)). This balanced approach offers flexibility for creative edits while preserving key spatial cues from the original scene.

\subsection{Content Harmonization} 
\label{sec:harmonization}
Our anchor extended attention steers the placement of the new content to align with the original scene. However, it does not guarantee precise pixel-level alignment. As can be seen in Fig. \ref{fig:extended_attn}(e), the legs of the horse are not perfectly aligned with the original video. To guarantee pixel-level alignment, a straightforward approach is to extract a mask of the new content $\bs{M}_{\text{VFX}}^{\text{rgb}}$ from the sampling with AnchorExtAttn (Eq. \ref{eq:AnchorExtAttn}) output $\mathcal{\hat{V}}_{\text{comp}}$. More concretely, we obtain this mask by applying a text-based segmentation model using the added object description provided by the VLM.
The mask can then be used to replace the pixels outside it with the corresponding pixels from the original video: $\mathcal{V}{\text{comp}}[\sim\bs{M}_{\text{VFX}}^{\text{rgb}}] = \mathcal{V}{\text{orig}}[\sim \bs{M}_{\text{VFX}}^{\text{rgb}}]$. While this preserves the unaffected regions, it often results in poor harmonization with the input video. 

To improve content harmonization, we propose repeating the sampling process with AnchorExtAttn (Eq. \ref{eq:AnchorExtAttn}) multiple times, progressively reducing the level of noise added at each step. This iterative approach gradually refines the new content's interaction with the original scene. 

As shown in Fig. \ref{fig:pipeline}, we update $\bs{x_{\text{comp}}}$ by adding a residual latent $\bs{x}_{\text{res}}=\bs{{M}_{\text{VFX}}\cdot(\hat{x}_{\text{comp}}-x_{\text{orig}}})$ to $\bs{x_{\text{orig}}}$, where  $\bs{{M}_{{\text{VFX}}}}$ is the latent representation of $\bs{{M}_{\text{VFX}}^{\text{rgb}}}$ (see Sec. \ref{app:latent_mask_extract} in SM for details on computing the latent $\bs{{M}_{\text{VFX}}}$). Notably, the addition of $\bs{x}_{\text{res}}$ is equivalent to directly replacing values of $\bs{x_{\text{orig}}}$ with those of $\bs{\hat{x}_{\text{comp}}}$ within $\bs{{M}_{\text{VFX}}}$. The final edited video is obtained by decoding their sum with the T2V model's VAE decoder. This allows each iteration to adjust the generated content's high-frequency details to better match the original video. We provide implementation details and summarize our method in Sec. \ref{app:imp_details} and Alg. \ref{alg:main} in the SM.



\begin{figure*}[htbp!]
    \centering
    \includegraphics[width=\textwidth]{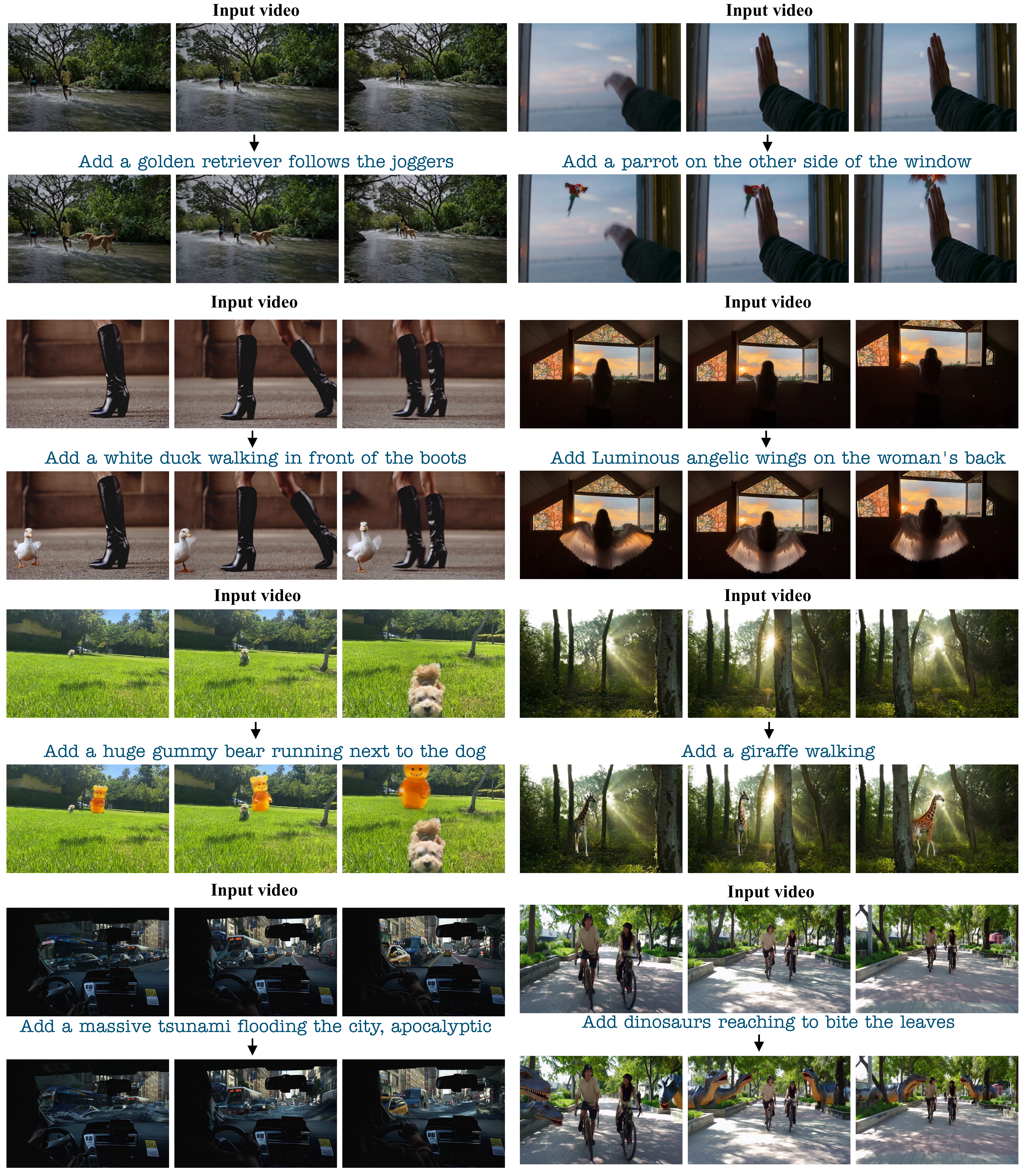}
    \vspace{-0.7cm}
    \caption{{\bf Sample Results of DynVFX.} Our method supports a wide range of scene augmentations across diverse scenarios while maintaining realistic interaction, occlusion, lighting, and camera motion, for example: a golden retriever consistent with camera movement, transparent wings revealing the woman’s silhouette at sunset, and a tsunami flooding the city yet realistically respecting the car dashboard. See SM for full videos.}
    \label{fig:results}
\end{figure*}

\begin{figure*}[ht!]
    \centering
    \includegraphics[width=\textwidth]{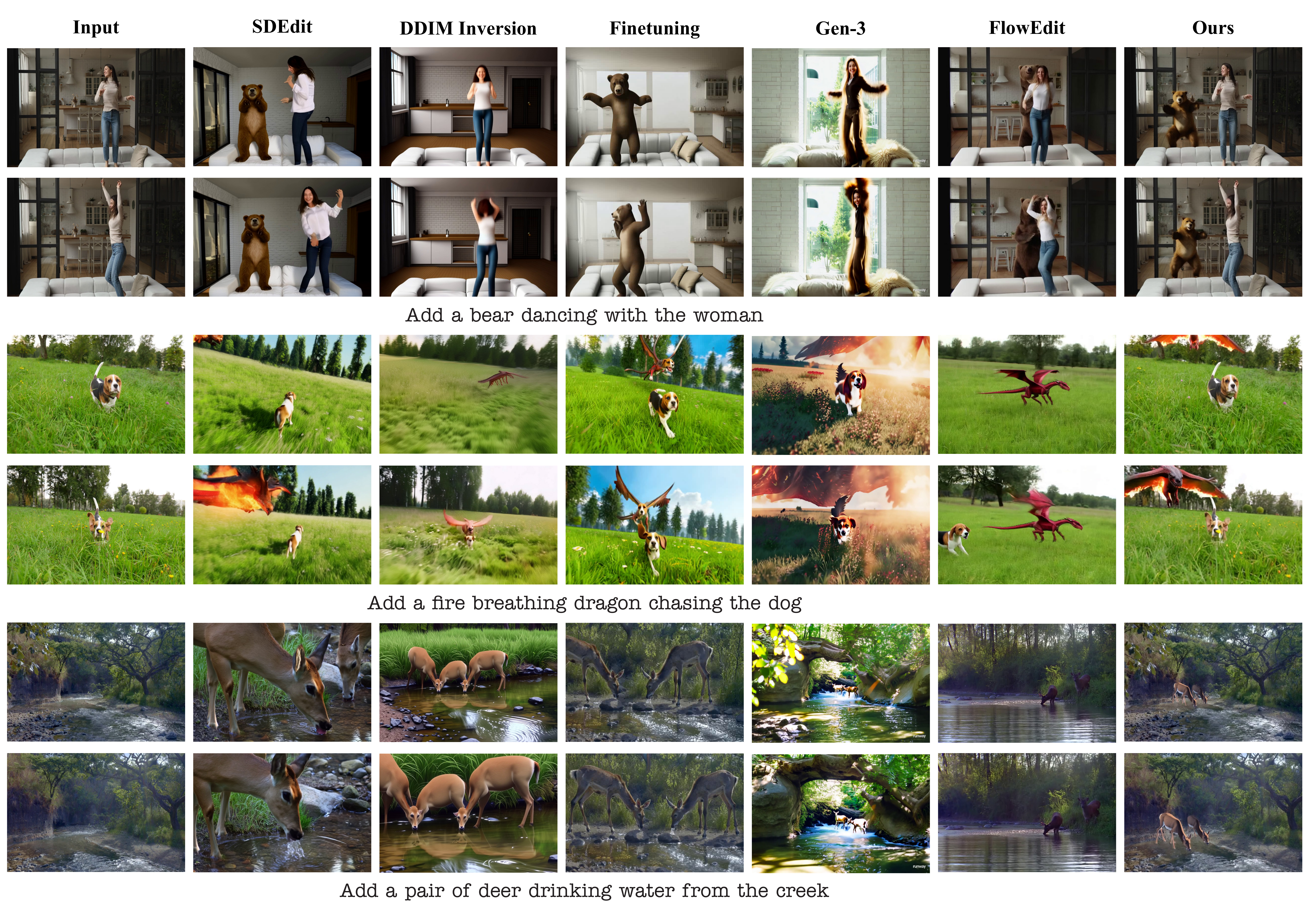}
    \vspace{-0.3cm}
    \caption{{\bf Qualitative Comparison of Text-Based Methods.} Sample results comparing our method to SDEdit \cite{meng2022sdedit}, DDIM inversion \cite{song2020_ddim}, Lora fine-tuning \cite{lora}, Gen-3 \cite{gen3} and FlowEdit \cite{kulikov2024flowedit}. As can be seen, our method better augments the original scene with new dynamic content that interacts naturally with existing elements in the scene. See SM for full video comparison.}
    \label{fig:comparison}\afterfigure
\end{figure*}

\section{Results}
\label{sec:results}
We evaluated our method over a set of 57 video-text edit pairs containing 36 unique publicly available videos from the web and the DAVIS  dataset \cite{davis}. These videos feature a wide range of complex scenes in terms of camera and object motion, lighting conditions, and physical environments. 
Our videos and implementation details are available in the SM.

Figs.~\ref{fig:teaser} and ~\ref{fig:results} show sample results of our method. As seen, our method facilitates natural integration of a broad range of visual effects, ranging from environmental effects (a tsunami in Fig. \ref{fig:ablation} and an explosion in Fig. \ref{fig:results}) to new object insertion (knight riding a horse in Fig. \ref{fig:results} and dancing bear in Fig. \ref{fig:comparison}).  In all examples, the new content is naturally localized in the scene, even in challenging scenarios involving multiple objects (dinosaurs or workers in Fig. \ref{fig:results}) and partial occlusions (a puppy in Fig. \ref{fig:teaser} and a giraffe in Fig. \ref{fig:results}). 

\begin{figure*} [ht!]
    \centering
\includegraphics[width=0.92\textwidth]{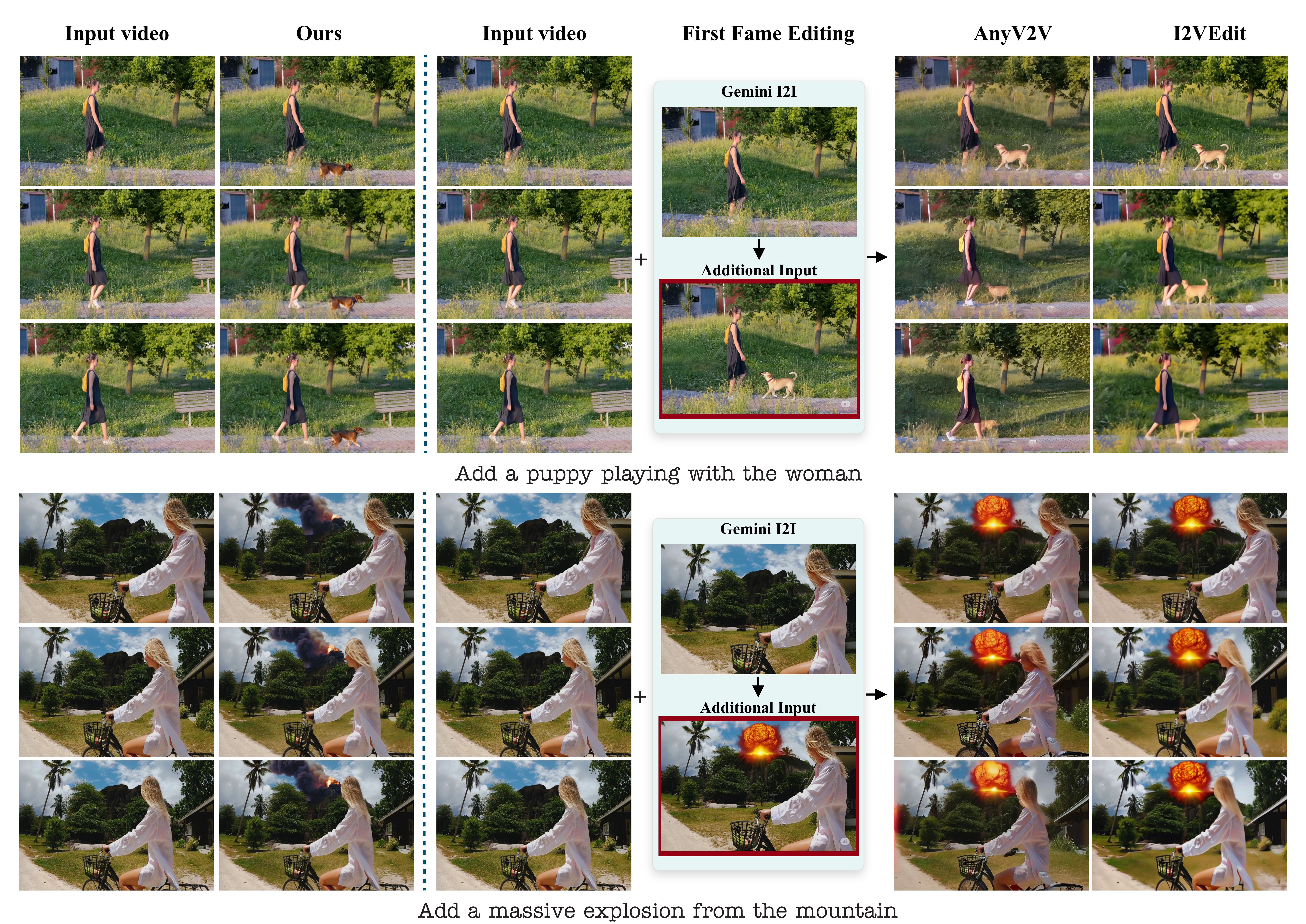}
\vspace{-0.1cm}
\caption{{\bf Comparison to Two-Stage Editing Methods.} Left pane: our results. Right pane: AnyV2V and I2VEdit. See SM for video comparisons.
}
\label{fig:comparisons_two_stage}
\end{figure*}

\subsection{Qualitative Evaluation}
We compare our method to the following baselines: \textit{(i)} SDEdit \cite{meng2022sdedit} using the same T2V model as ours, \textit{(ii)} DDIM inversion \cite{song2020_ddim} and sampling with the target prompt, \textit{(iii)} LORA fine-tuning \cite{lora} of the T2V model and sampling with the target prompt, \textit{(iv)} Gen-3 \cite{gen3} video-to-video model, designed for video stylization, \textit{(v)} FlowEdit \cite{kulikov2024flowedit} and two-stage editing approaches: \textit{(vi)} AnyV2V \cite{ku2024anyv2v} and \textit{(vii)} I2VEdit \cite{ouyang2024i2vedit}. 

Figs. \ref{fig:comparison} and \ref{fig:comparisons_two_stage} show a qualitative comparison with the baselines. As can be seen, all baselines exhibit different limitations in maintaining scene fidelity while introducing new content. SDEdit \cite{meng2022sdedit} and FlowEdit \cite{kulikov2024flowedit} both manage to fulfill the edit prompt, yet the scene may significantly deviate from the original scene in terms of appearance, motion, positioning, or scale (e.g., deer in the creek). DDIM inversion is not suitable for editing. LORA fine-tuning suffers from the trade-off between preserving aspects of the original scene and adding new content to the scene: either overfitting original scene appearance (e.g., adding a dragon-dog hybrid), or underfitting the original layout (e.g., incorrect scale of deer).
Gen-3 is conditioned on structure and content representation extracted from the input video, hence it tends to significantly alter scene appearance and does not allow the insertion of new objects that change the scene layout. Both AnyV2V and I2VEdit take the first edited frame as additional input, but still struggle to maintain the coherent motion of the generated objects. In particular, AnyV2V often fails to accurately propagate the edits through the video, leading to distorted geometries. In each case, these limitations affect the overall scene coherence and realism of the added elements. Our method successfully adds new content to the scene, achieving high fidelity to the user instructions while allowing for natural interactions between original and added elements  (e.g., natural interaction between woman and bear). 
We provide details about the baselines comparison in Sec. \ref{app:base_comp_details} in the SM.
Additionally, we provide qualitative comparisons with VFX reference-based methods \cite{pika, kling} in Sec. \ref{app:addition_qual_compare} and Fig. \ref{fig:comparison_pika} in the SM.

\begin{table*}[t]
\caption{{\bf Quantitative Comparison.} We compare our method with other video editing methods and ablations using metrics for edit fidelity (CLIP Directional) and original content preservation (masked SSIM), a VLM-based evaluation of text alignment, visual quality, harmonization, and motion realism, and a user study on content integration and edit harmonization. User study values indicate the percentage of participants who preferred our method over each baseline.}
\centering
\small
\setlength{\tabcolsep}{4pt}
\begin{tabular}{l|cc|cccc|cc}
\toprule
Method & \multicolumn{2}{c|}{Metrics} & \multicolumn{4}{c|}{VLM-based evaluation} & \multicolumn{2}{c}{User Study} \\
\midrule
 & \makecell{CLIP\\Directional} & SSIM & \makecell{Text\\Alignment} & \makecell{Visual\\Quality} & \makecell{Edit\\Harmonization} & \makecell{Dynamics\\Score} & \makecell{Content\\Integration} & \makecell{Edit\\Harmonization} \\
\midrule
Gen-3 & 0.142 & 0.283 & 0.410 & 0.586 & 0.363 & 0.374 & 98.687 & 94.691 \\
LORA finetuning & 0.292 & 0.349 & 0.781 & 0.755 & 0.726 & 0.714 & 92.221 & 71.968 \\
DDIM inv. sampling & 0.193 & 0.440 & 0.465 & 0.609 & 0.448 & 0.448 & 98.951 & 93.145 \\
SDEdit (0.9) & 0.290 & 0.321 & 0.778 & 0.750 & 0.717 & 0.730 & 99.154 & 75.645 \\
SDEdit (0.6) & 0.105 & 0.558 & 0.381 & 0.703 & 0.406 & 0.390 & 92.655 & 92.470 \\
FlowEdit & 0.288 & 0.414 & 0.771 & 0.765 & 0.712 & 0.736 & 97.346 & 65.494 \\
AnyV2V & 0.300 & 0.468 & 0.678 & 0.669 & 0.615 & 0.619 & 91.142 & 77.747 \\
I2VEdit & 0.288 & 0.524 & 0.769 & 0.773 & 0.724 & 0.719 & 91.030 & 82.563 \\
\midrule
w/o AnchorExtAttn & \textbf{0.325} & 0.691 & 0.765 & 0.713 & 0.670 & 0.687 & 81.417 & 81.109 \\
w/o Iterative Refinement & 0.291 & 0.736 & 0.749 & 0.741 & 0.697 & 0.696 & 80.122 & 80.492 \\
w/o VLM Protocol & 0.254 & 0.741 & 0.705 & 0.755 & 0.669 & 0.667 & 71.360 & 73.708 \\
Ours & 0.307 & \textbf{0.749} & \textbf{0.843} & \textbf{0.784} & \textbf{0.778} & \textbf{0.775} & - & - \\
\bottomrule
\end{tabular}
\label{tab:performance}
\end{table*}
\subsection{Quantitative Evaluation} \label{sec:quant}
We numerically evaluate our results using the following metrics:

\noindent \textit{(i) Edit fidelity.}  Following previous works (e.g., \cite{hsu2024autovfx,Tewel2024AdditTO}), we measure per-frame Directional CLIP Similarity \cite{clip,gal2021stylegannada} to assess the alignment between the change in the source and the target prompt, and the change between the source and edited frames. 

\noindent \textit{(ii) Original content preservation.} We evaluate how well the edited video preserves the original content outside the modified region. To this end, we segment the new content in the edited video using \cite{zhang2024evfsamearlyvisionlanguagefusion}, and compute the masked Structural Similarity Index (SSIM) over the regions complementary to the edited ones.

\noindent \textit{(iii) \emph{VLM quality evaluation.}}  Inspired by \cite{hsu2024autovfx, huang2023vbench}, we employ a VLM to expand the per-frame metrics above as follows. We input several frames from the edited videos into the VLM and instruct it to evaluate four key aspects: how well the edit follows the text prompt (\emph{Text Alignment}), the overall visual quality of the edited frames (\emph{Visual Quality}), how well the new content is harmonized with the source frames (\emph{Edit Harmonization}), and the realism of the added object's dynamics relative to the scene  (\emph{Dynamics Score}). For each aspect, the VLM outputs a score between 0 and 1, with higher scores indicating better performance. In the SM, we include our evaluation protocol and the justification of its use as an evaluation metric, showing its alignment with human preferences and consistent outcomes with an alternative VLM backbone. We include this in Sec. \ref{app:vlm_eval_metric} and Table \ref{tab:performance_3} in the SM.

Table \ref{tab:performance} and Fig. \ref{fig:metrics} in the SM present the results of the described metrics on a set of 57 video-text edit pairs comprising 34 unique videos. As shown, our method outperforms the baselines in both SSIM and Directional CLIP metrics, demonstrating superior edit fidelity while maintaining higher structural similarity in the unedited regions. The VLM-based evaluation aligns with this assessment and further  shows that our method produces videos that achieve better content integration and greater motion realism.

\noindent \textit{(iv) User study.} We conducted a user study to evaluate the ability to integrate new content while preserving the original video. Participants were shown the input video, a text description of the new content, our result, and a baseline output. They were asked two questions: ``Which video better preserves the original footage while adding new content?'' and ``Which video better integrates the new content in a realistic and seamless way?''. In total, we collected 17,160 user judgments from 330 users. As seen in Table~\ref{tab:performance}, our method is consistently preferred over all baselines.

\subsection{Ablations}
We ablate key design choices of our method: Anchor Extended Attention, iterative updates of the edit, and our VLM as a VFX Assistant protocol, by excluding each component from our framework. The ablation of our VLM protocol can be found in Sec. \ref{app:ablate_vfx_protocol} and Fig. \ref{fig:ablate_protocol} in the SM.

As seen in Fig.~\ref{fig:ablation}(c),   omitting AnchorExtAttn leads to new content being misaligned relative to the original scene, with the added content poorly integrated into the original scene. 
Applying only the first iteration of our method (w/o iterative refinement, Fig.~\ref{fig:ablation}(d)) results in a better alignment with the input video, but the composition is still unstable, as evident, for example, in the puppy's body hovering over the box in the first scene. 
Our full method achieves better composition with proper spatial relationships, demonstrating the importance of both components for realistic scene editing (Fig.~\ref{fig:ablation}(f)). We numerically evaluate each ablation with the same set of metrics described in Sec~\ref{sec:quant} and report them in Table~\ref{tab:performance}. An additional ablation example can be found in Fig. \ref{fig:ablation_extra} in the SM.

\begin{figure}[htbp]
    \centering
    \includegraphics[width=0.95\linewidth]{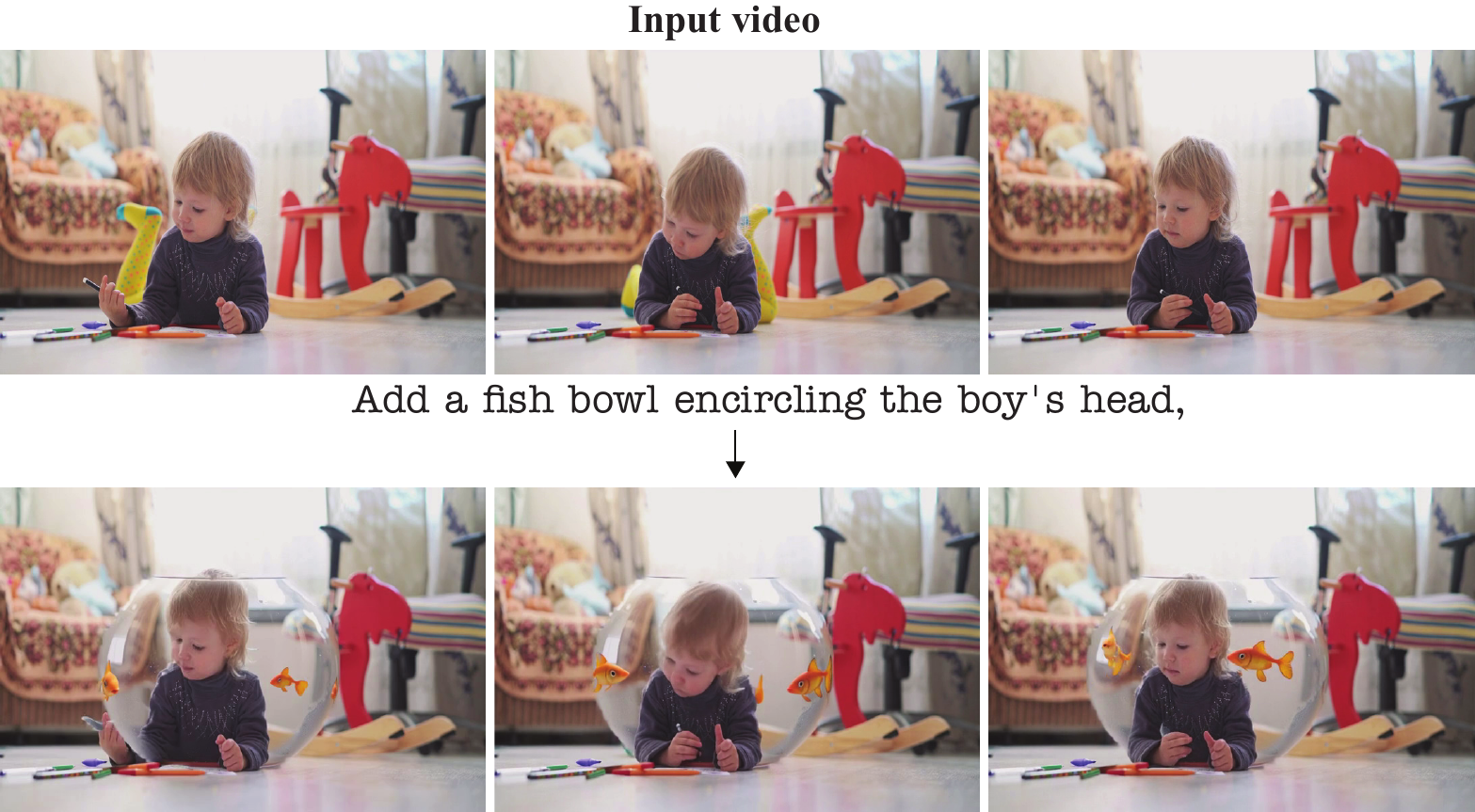}
    \caption{{\bf Limitations.} In some cases, the T2V model struggles to precisely follow the edit prompt.}
    \label{fig:limitation}\afterfigure 
\end{figure}

\section{Discussion and Conclusions}
\label{sec:conclusions}
We introduced the task of augmenting real videos with new dynamic content based on a user-provided instruction. We presented a zero-shot method utilizing the T2V model in a feature manipulation framework, enabling correct localization and natural blending of new content with existing video elements. 

As our method is built upon the pre-trained T2V model, the quality of the generated edits is inherently tied to the performance and capabilities of the underlying model. As shown in Fig. \ref{fig:limitation}, the T2V model sometimes struggles to generate videos that precisely follow the edit prompt; additional failure cases and limitations are detailed in Sec. \ref{app:limitations} and Fig. \ref{fig:limitation_2} in the SM.
While the quality of the edits is bound by the capabilities of the T2V model, our method is model-agnostic and can be re-implemented with future transformer-based T2V models. We anticipate its performance will improve as more powerful models become available.
Even under these constraints, our method significantly outperforms baselines, expanding the capabilities of pre-trained text-to-video
diffusion models.



\newpage

\bibliographystyle{ACM-Reference-Format}
\bibliography{bibliography}

\newpage


\appendix
\section*{\Large\textbf{Appendix}\\}


\setcounter{tocdepth}{2} 
\begin{center}
  \textsc{Table of Contents}
\end{center}
\startcontents[app]             
\setcounter{tocdepth}{2}
\printcontents[app]{l}{1}{}


\section{Implementation Details}\label{app:imp_details}
\subsection{Models}\label{app:models} 
\myparagraph{Text-to-Video Model.} We use a publicly available CogVideoX-5B \cite{hong2022cogvideo,yang2024cogvideox} text-to-video model, which can generate videos with up to $480 \times 720$ pixels resolution, 6 seconds in length, 49 frames at 8 fps. This model is a transformer-based model that processes both text and video modalities together. 

\vspace{-0.1cm}
\myparagraph{Segmentation Model.} To segment the prominent objects in the video and the newly generated content, we utilize EVF-SAM2 \cite{zhang2024evfsamearlyvisionlanguagefusion} - a text-based video segmentation model based on SAM2 \cite{Ravi2024SAM2S}.

\vspace{-0.1cm}
\myparagraph{Visual Language Model.} Our vision-language model of choice is GPT-4o \cite{achiam2023gpt4}, which we use through the official OpenAI API.

\begin{figure}[h!]
    \centering
 \includegraphics[width=\columnwidth]{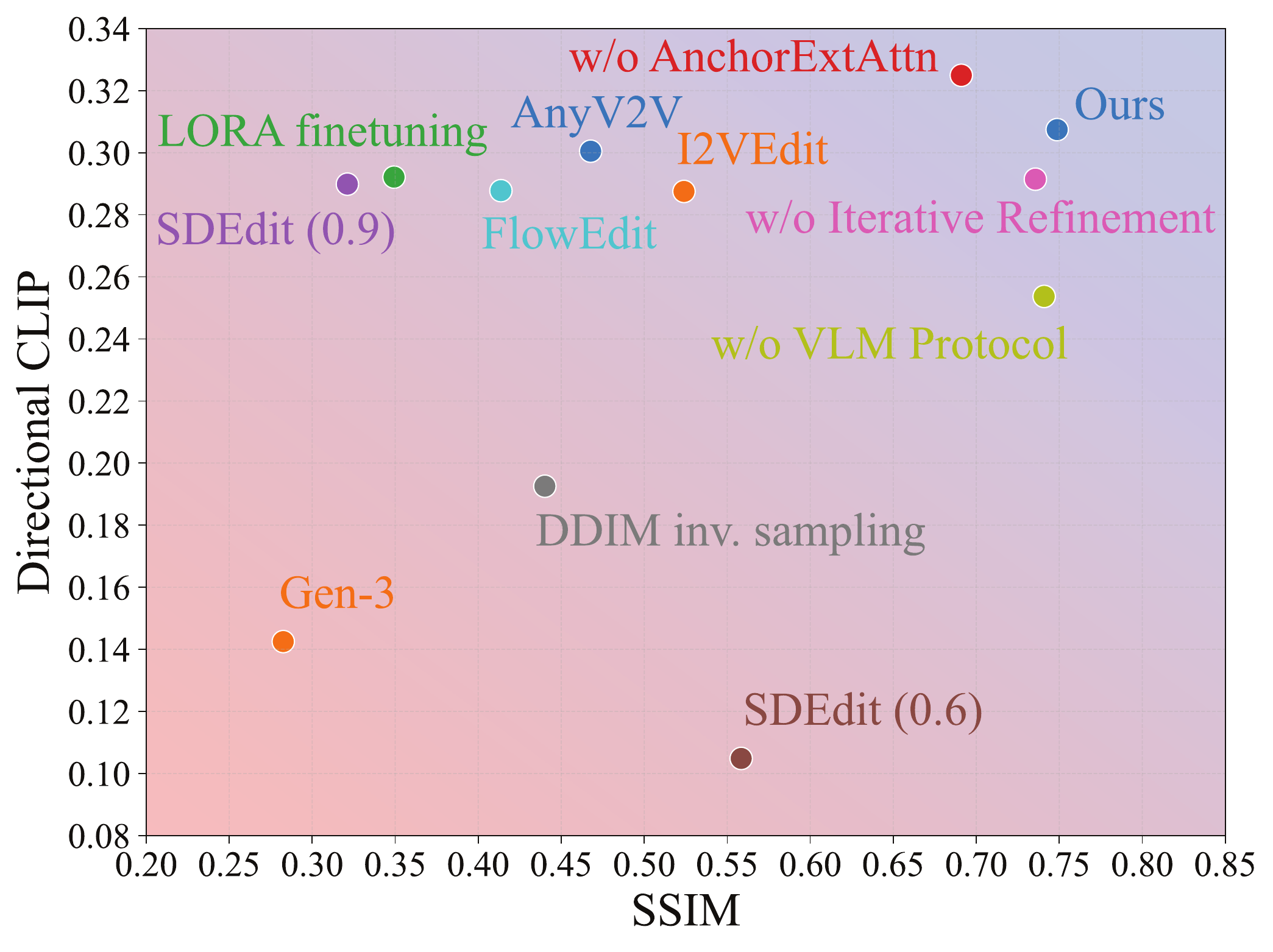} 
            \vspace{-0.7cm}
    \caption{{\bf Metrics Visualization.} We measure CLIP Directional score (higher is better) and masked SSIM (higher is better). Our method demonstrates a better balance between these two metrics. See Sec. \ref{sec:quant} for full discussion.}
  \label{fig:metrics}\vspace{-3mm}
\end{figure}

{\small 
\begin{algorithm}[htbp!]
\small
\begin{flushleft}
    \caption{\bf{DynVFX Algorithm}}
    \label{alg:main}
    \textbf{Input:}
    \begin{algorithmic}
     \State $\mathcal{V}_{\text{orig}}$, $\mathcal{P}_{\text{VFX}}$ \hfill $\triangleright$ Input video \& instruction prompt
     \State  $\mathcal{\tau}_{\text{A}}$ \hfill $\triangleright$ Extended Attention threshold
     \State  ${\Psi}$ \hfill $\triangleright$ Video segmentation model
     \State  ${\text{VLM}}$ \hfill $\triangleright$ Vision Language model 
    \end{algorithmic} 
    \textbf{Preprocess:}
    \begin{algorithmic}

    \State $\mathcal{P}_{\text{comp}}  \gets \text{VLM} 
    [\mathcal{V}_{\text{orig}}, \mathcal{P}_{\text{VFX}}]$ \hfill $\triangleright$ Composition Prompt
    \State $\mathcal{O}_{\text{orig}}, \mathcal{O}_{\text{edit}}   \gets \text{VLM}
    [\mathcal{V}_{\text{orig}}, \mathcal{P}_{\text{VFX}}]$ \hfill $\triangleright$ Original objects and VFX object
    
    \State ${M}_{orig} \gets \texttt{Get-Latent-Mask}(\Psi;\mathcal{V}_{\text{orig}}, \mathcal{O}_{\text{orig}})$ \hfill $\triangleright$ Extract source masks
    \State ${x}_{orig} \gets \texttt{Encode}[\mathcal{V}_{\text{orig}}]$ \hfill $\triangleright$ {\text{Encode video into latent space}}

    \State $\mathbf{K}_\text{orig}, \mathbf{V}_\text{orig} \gets \text{DDIM-Inv}[{x}_{orig}] \quad \forall t\in[T]$
    \end{algorithmic}
    \textbf{For} $t=\tilde{T},\dots,T_{min}$ \textbf{do}
    \begin{algorithmic}
        \State $\bs{x}_{res} = \bs{0}$ \hfill $\triangleright$ initialize the residual latent
        \State $\bs{x}_{comp} = {x}_{orig} + {x}_{res}$ 

        \State $\bold{\text{if}} \ t>\mathcal{\tau}_{A} \ \bold{\text{then}} \ {K^E,V^E} \gets \mathcal{F}(K_\text{orig} , M_\text{orig}), \mathcal{F}(V_\text{orig} , M_\text{orig})$
        \\
        $\bold{\text{else  }} 
        {K^E,V^E} \gets \emptyset $ \hfill
        \State $\bs{\hat{x}}_{comp} \gets \texttt{Sampling}[{x}_{comp}, \mathcal{P}_{\text{comp}}, t; \texttt{AnchorExtAttn}{[K^E,V^E]}]$ \hfill 
        \State $\mathcal{\hat{V}}_{comp} \gets \texttt{Decode}(\bs{\hat{x}}_{comp})$ \hfill $\triangleright$ Decode latent
        \State $\bs{M}_{VFX} \gets \texttt{Get-Latent-Mask}(\Psi;\mathcal{\hat{V}}_{comp}, \mathcal{O}_{\text{edit}})$ \hfill $\triangleright$ Extract VFX masks
        
        \State $\bs{x}_{res} = \bs{M}_{VFX} \cdot ({\hat{x}}_{comp} - {x}_{orig})$ \hfill
    \end{algorithmic}
    $\bs{x}_{comp} = {x}_{orig} + {x}_{res}$ \\
    ${\mathcal{V}}_{\text{comp}} \gets \texttt{Decode}[{x}_{comp}]$ \hfill $\triangleright$ {\text{Output video}}\\
    \textbf{Output:} $\mathcal{V}_{\text{comp}}$ 
\end{flushleft}\vspace{-0.1cm}
\end{algorithm}}
\subsection{Keys and Values Extraction}\label{app:key_value_extract} 
Following \cite{Tumanyan_2023_CVPR,yatim2023spacetime}, to obtain the T2V diffusion model intermediate latents, we apply DDIM inversion (applying DDIM sampling in reverse order) on the input video. We perform inversion solely to extract the keys and values for AnchorExtAttn, while the sampling itself is initialized from a noisy $x_{comp}$. 
For the inversion process, we use 250 steps, with an empty string as the text prompt. We derive keys and values from the noisy latents produced by the inversion process (not from the reconstruction).
During the sampling iterations in our method, we use these extracted keys and values for AnchorExtAttn. (Eq. \ref{eq:AnchorExtAttn}) randomly samples a selection of anchor points, which are re-selected in each layer of the model.
For all our results, we fix the hyperparameters and only adjust the dropout percentage in the AnchorExtAttn: 5\textit{\%} dropout for background and 70\% for foreground in regular edits, while global edits use 10\% for the foreground.

\subsection{Obtaining $M_{orig}$}\label{app:obtain_morig} 
\begin{figure}[htbp!]
\centering
\includegraphics[width=\columnwidth]{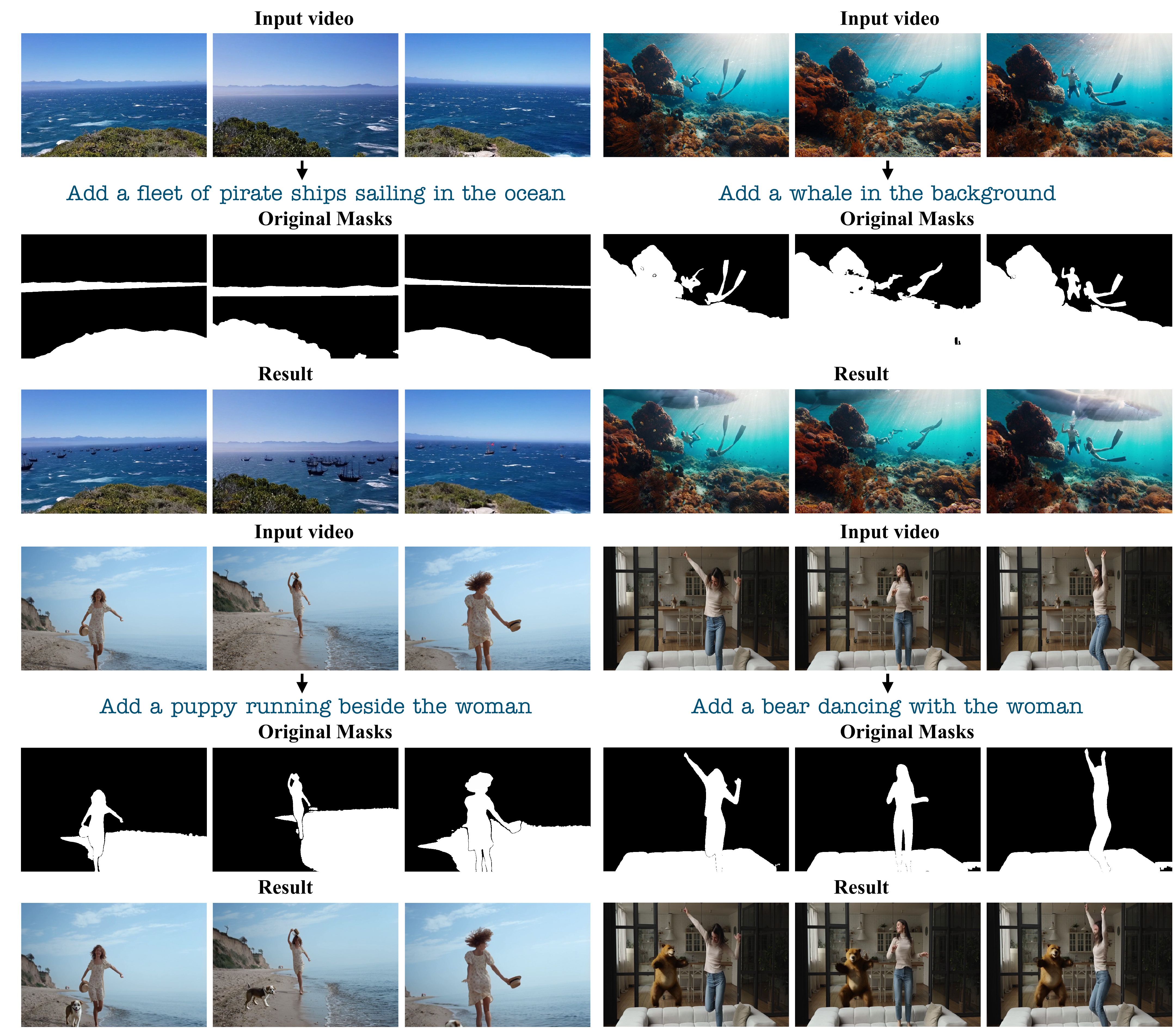}
    \vspace{-0.7cm}
    \caption{\looseness-1 {\bf Qualitative Examples of $M_{orig}$.} Each example includes the input video, user instruction, original masks, and corresponding result. In the top-right whale edit, our VLM protocol identifies the prominent elements to be coral and divers. In the top left pirate ships edit, the prominent elements are the cliffs. In the bottom left puppy edit, the woman and the ocean are identified. And in the bottom right bear edit, the woman and the couch are chosen.}
    \afterfigure
    \label{fig:orig_masks}
\end{figure}
To get $M_{\text{orig}}$, we ask a VLM  to provide a list of foreground objects in the original video $\mathcal{O}_{\text{orig}}$ (typically the most spatially prominent elements within a scene) and then we obtain corresponding masks using EVF-SAM \cite{zhang2024evfsamearlyvisionlanguagefusion} - a text-based segmentation model. 

\begin{figure*}[t!]
    \centering
\includegraphics[width=\textwidth]{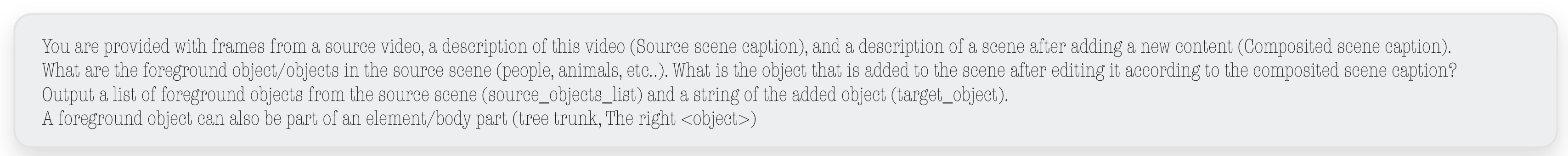}
    \caption{{\bf System Prompt For Obtaining $\mathcal{O}_{\text{orig}}$ and $\mathcal{O}_{\text{edit}}$.} Given the user prompt, the input video's key frames and $\mathcal{P}_{\text{comp}}$, the VLM identifies prominent elements from the original scene $\mathcal{O}_{\text{orig}}$ and the new content to be added $\mathcal{O}_{\text{edit}}$. These outputs are used in our framework for the application of SAM, used to aid localization and harmonization of the new content.}
    \label{fig:mask_system_prompt}
\end{figure*}

Specifically, given the user instruction, the input video’s key frames and $\mathcal{P}_{\text{comp}}$, our VLM protocol identifies the prominent elements from the original scene using the system prompt in Fig.~\ref {fig:mask_system_prompt}.
Next, each element (represented by words) is fed to EVF-SAM to produce masks, and their union forms $M_{orig}$, which is used in our AnchorExtAttn.
Examples of such masks can be seen in Fig.~\ref{fig:orig_masks}.

\subsection{Latent Mask Extraction}\label{app:latent_mask_extract} 
As discussed in Sec. \ref{sec:harmonization}, we iteratively update the residual latent $\bs{x}_{res}$ in the regions where the new content appears. This requires calculating the mask of the new content in the latent space. To do this, we first apply the segmentation model \cite{zhang2024evfsamearlyvisionlanguagefusion} to the current output of SDEdit and get the mask of the new content in RGB space.  However, the VAE in the T2V diffusion model involves both spatial and temporal downsampling, making it challenging to directly map RGB pixels to their corresponding latent regions. To address this, we encode the RGB masks through the VAE-Encoder and apply clustering to partition the resulting latents into two groups, effectively producing downsampled masks that align with the latent space representation. 

\subsection{Runtime and Memory Usage}\label{app:runtime_memory} 
Our method's most computationally intensive parts are - DDIM inversion, which takes \mytextapprox 15 minutes, and iterative updates of the edit residual, which takes \mytextapprox 20 minutes, while a single iteration of sampling with AnchoExtAttn takes between 2-5 minutes (depending on the noise level). Importantly, DDIM inversion needs to be performed only once per video and can support multiple subsequent edits, making the process more efficient when applying various modifications to the same video content. Querying the VLM and EVF-SAM adds negligible runtime (\mytextapprox 5 and \mytextapprox 20 seconds correspondingly).
We run our method on a NVIDIA A100, with 72GB memory in usage.



\section{Baselines Comparison Details}\label{app:base_comp_details}  
The Baseline runtimes are \textit{(i)} SDEdit - \mytextapprox 5 minutes \textit{(ii)} DDIM inversion + Sampling - \mytextapprox 20 minutes \textit{(iii)} LORA fine-tuning - \mytextapprox 2 hours, \textit{(iv)} Gen-3 - \mytextapprox 30 seconds, \textit{(v)} FlowEdit - \mytextapprox 3 minutes, \textit{(vi)} AnyV2V - \mytextapprox 3 minutes and \textit{(vii)} I2VEdit - \mytextapprox 30 minutes.

In terms of utilized Video Models, baselines \textit{(i)} SDEdit, \textit{(ii)} DDIM-inversion+Sampling, and \textit{(iii)} LORA fine-tuning utilized the same text-to-video model \cite{yang2024cogvideox} as our method, \textit{(iv)} Gen-3 utilizes a proprietary video-to-video model by Runway (Alpha model via the publicly accessible web-based API), \textit{(v)} FlowEdit is applied with Hunyuan Video model \cite{kong2024hunyuanvideo}, \textit{(vi)} AnyV2V uses I2VGen-XL model \cite{2023i2vgenxl}, supporting 15 frames, and \textit{(vii)} I2VEdit utilizes Stable Video Diffusion model \cite{blattmann2023stablevideodiffusionscaling} supporting 27 frames. For AnyV2V and I2VEdit, the first frames were edited with Gemini image editing \cite{gemini}.

For LORA fine-tuning baseline, we use the following default hyperparameters: Adam optimizer \cite{adam}, $1e-4$ learning rank, LORA rank 128, 800 fine-tuning steps. For comparison with Gen-3 \cite{gen3}, we set the "Structure Transformation" hyperparameter to 5.

For FlowEdit, we use source and target prompts produced from our VLM-VFX protocol, as this method heavily relies on source and target prompting.


\section{Limitations}\label{app:limitations} 
    \vspace{-0.5cm}
\begin{figure}[htbp!]
    \centering    
    \includegraphics[width=\linewidth]{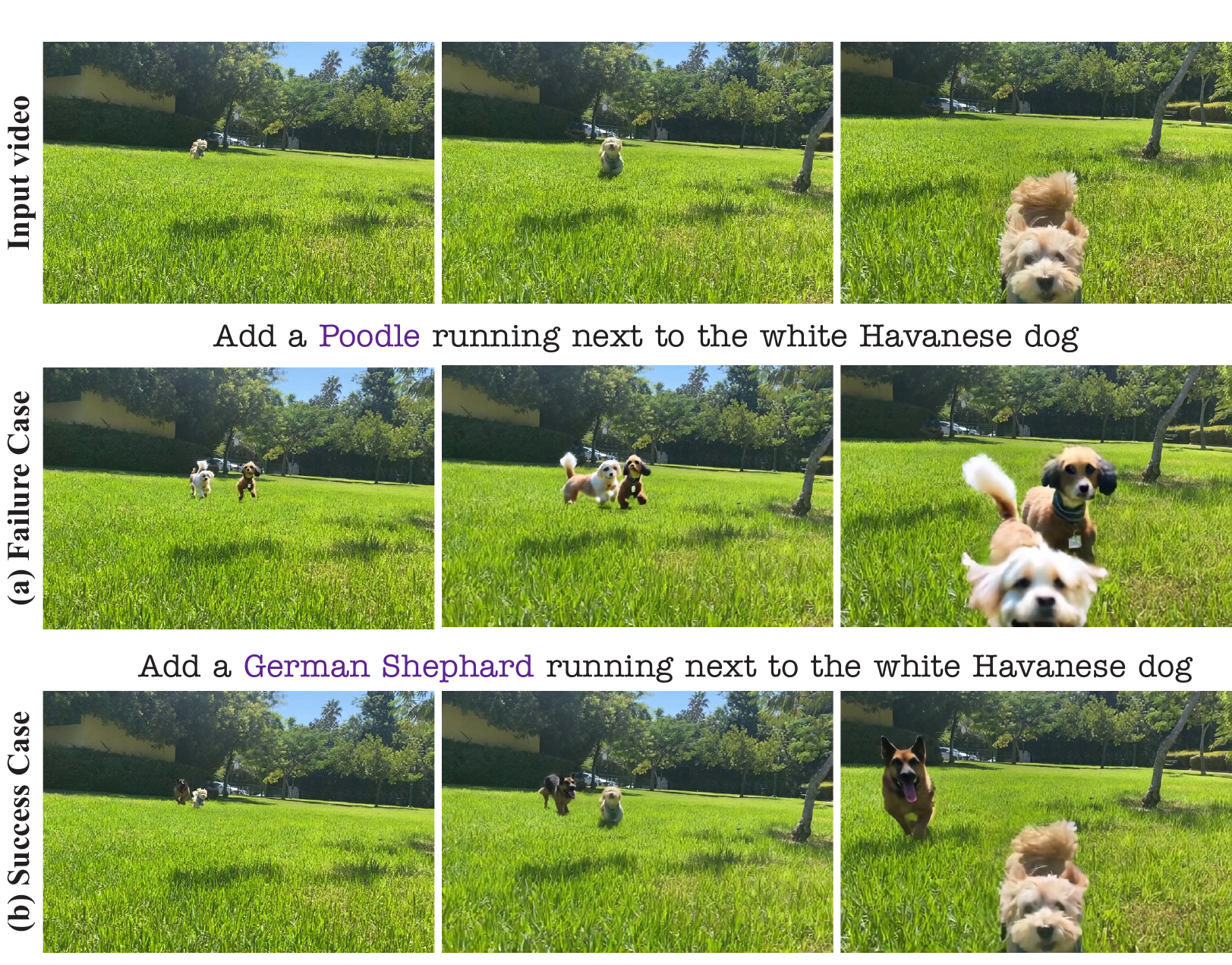}
    \vspace{-0.7cm}
    \caption{{\bf Failure Case Example.} When trying to add content that already exists in the scene (e.g., an additional dog), it is best to specify the differences between them in the user instructions. As can be seen, when trying to add a Poodle (similar appearance and size to a Havanese), EVF-SAM fails to distinguish between the two. However, when trying to add a German Shepherd, the segmentation model manages to make the distinction.}\afterfigure
    \label{fig:limitation_2}
\end{figure}
\begin{figure*}[h!]
    \centering
 \includegraphics[width=0.93\linewidth]{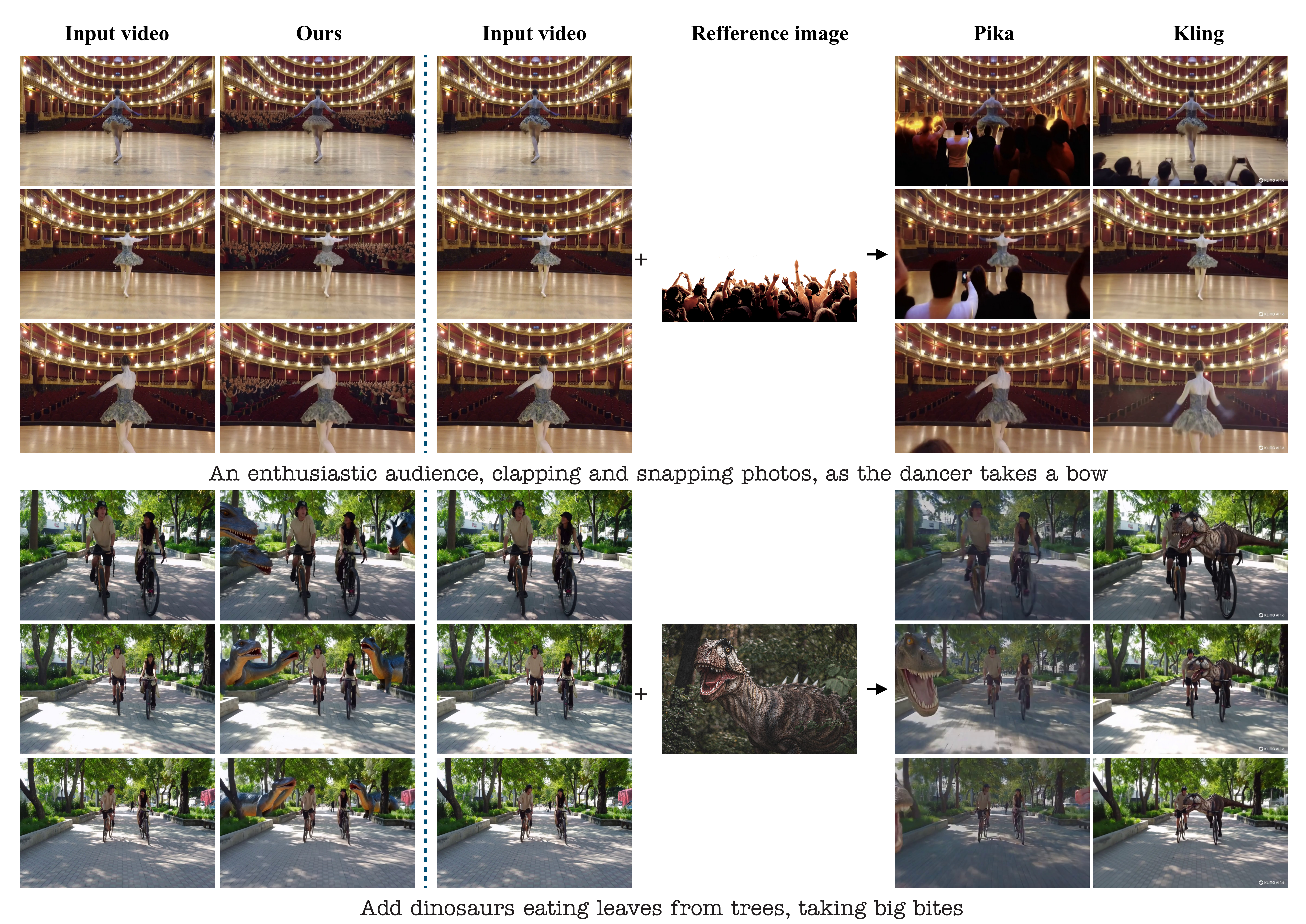}\vspace{-0.3cm}
    \caption{{\bf Comparison to Reference-Based Insertion Methods.} The left plane shows the results of our method, while the right plane shows the results of Pika and Kling. As can be seen, these reference-based methods struggle to generate global effects.}
\label{fig:comparison_pika}\vspace{-3mm}
\end{figure*}
\begin{figure}[htbp]
    \centering
\includegraphics[width=0.9\columnwidth]{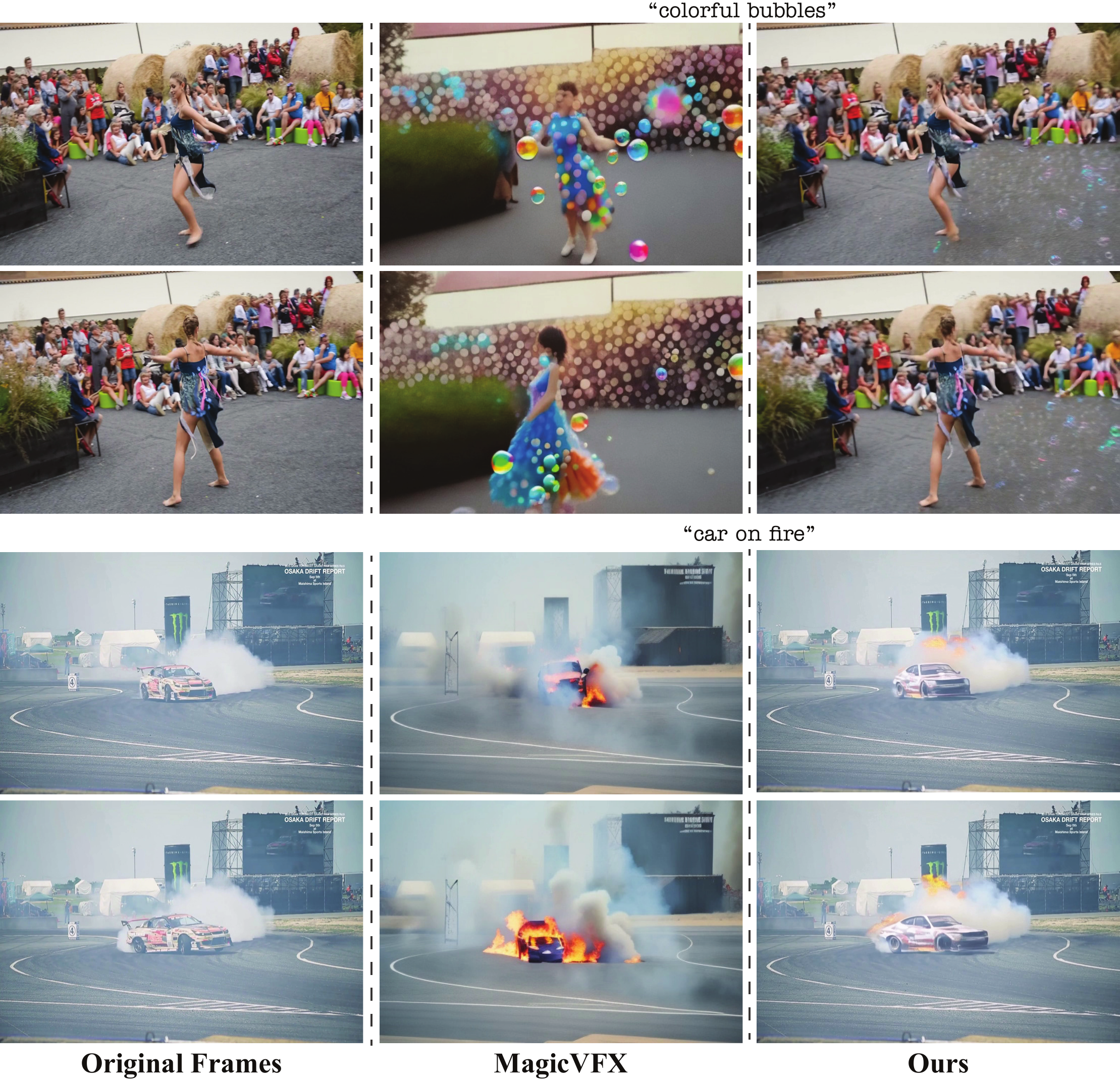}
     \vspace{-0.3cm}
    \caption{{\bf Comparison to MagicVFX.} MagicVFX (second column) deviates significantly from the original video. As seen in the third column, our method successfully adds new content while maintaining high fidelity to the original scene.}
    \label{fig:magicvfx_comparison}\vspace{-3mm}
\end{figure}
Our method relies on the target segmentation provided by a text-based segmentation model \cite{zhang2024evfsamearlyvisionlanguagefusion}. While this solution often succeeds for most edits, it can sometimes produce inaccurate masks and fail to account for effects like shadows and reflections if not specified in the text prompt. To handle such cases, users can add a specification in text to include these secondary effects. Furthermore, the segmentation model may struggle in scenarios where multiple entities with similar visual characteristics are present in the scene. As shown in Fig. ~\ref{fig:limitation_2}, segmentation issues can occur when adding content of similar appearance and size. However, when the entities are distinguishable enough, the new content is successfully integrated into the scene.

Furthermore, our framework performs a single-shot edit and does not support interactive, in-context refinements - once an edit is applied, users cannot request follow-up adjustments (e.g., tweaking object placement or appearance) without restarting the process. While VFX pipelines typically involve several feedback-driven iterations to perfect a scene, our current design does not support such incremental revisions. Future work could integrate an interactive editing loop, potentially via an autoregressive T2V model, to allow users to make successive edits without reinitializing the entire workflow.

\begin{figure*}[ht!]
    \centering
    \includegraphics[width=0.93\linewidth]{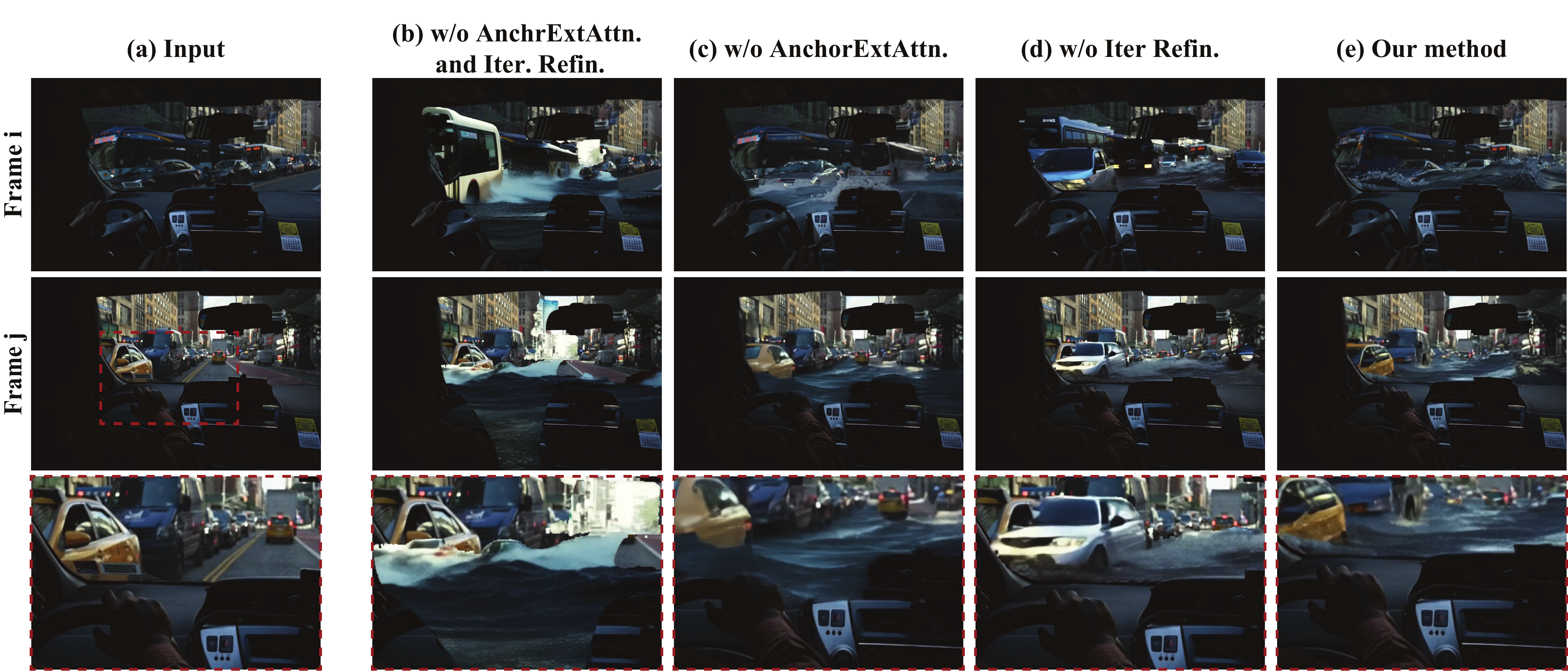} 
    \vspace{-3mm}
\caption{{\bf Additional Example for Ablations.} (b) Excluding both AnchorExtAttn and the Iterative refinement process results in significant misalignment with the original scene and poor harmonization (e.g., the size of the puppy relative to the scene and boundary artifacts). (c) Omitting AnchorExtAttn leads to incorrect positioning of the new content.  (d) Removing iterative refinement results in poor harmonization. Our full method (e) exhibits good localization and harmonization of the edit.}
\label{fig:ablation_extra}\vspace{-3mm}
\end{figure*}

\section{Additional Qualitative Comparisons}\label{app:addition_qual_compare} 
We perform additional qualitative comparisons to MagicVFX \cite{Guo2024MagicVFXVE} and two reference-based methods: Pikadditions \cite{pika} and Kling \cite{kling}. As can be seen in Fig.~\ref{fig:magicvfx_comparison}, MagicVFX struggles to remain faithful to the original scene and has lower visual quality compared to our method.  Furthermore, as illustrated in Fig. \ref{fig:comparison_pika}, both Pikadditions and Kling struggle to generate global effects. As can be seen, the audience members are incorrectly overlaid in front of the dancer. Additionally, the dinosaur occludes one of the bikers, and only one dinosaur is generated, although we prompt for a global edit. This demonstrates scenarios where text-based insertion is preferable over reference-based insertion. Even though such methods do not explicitly require the control of location and dynamics of the new content (as our method), they rely on the reference assets, which limits their applicability whenever a suitable example cannot be found for the given video.




\section{VLM Prompting}\label{app:vlm_prompting}
\subsection{Our VLM as a VFX Assistant Protocol}\label{app:our_vfx_protocol} 
\begin{figure*}[htbp]
    \centering
\includegraphics[width=\textwidth]{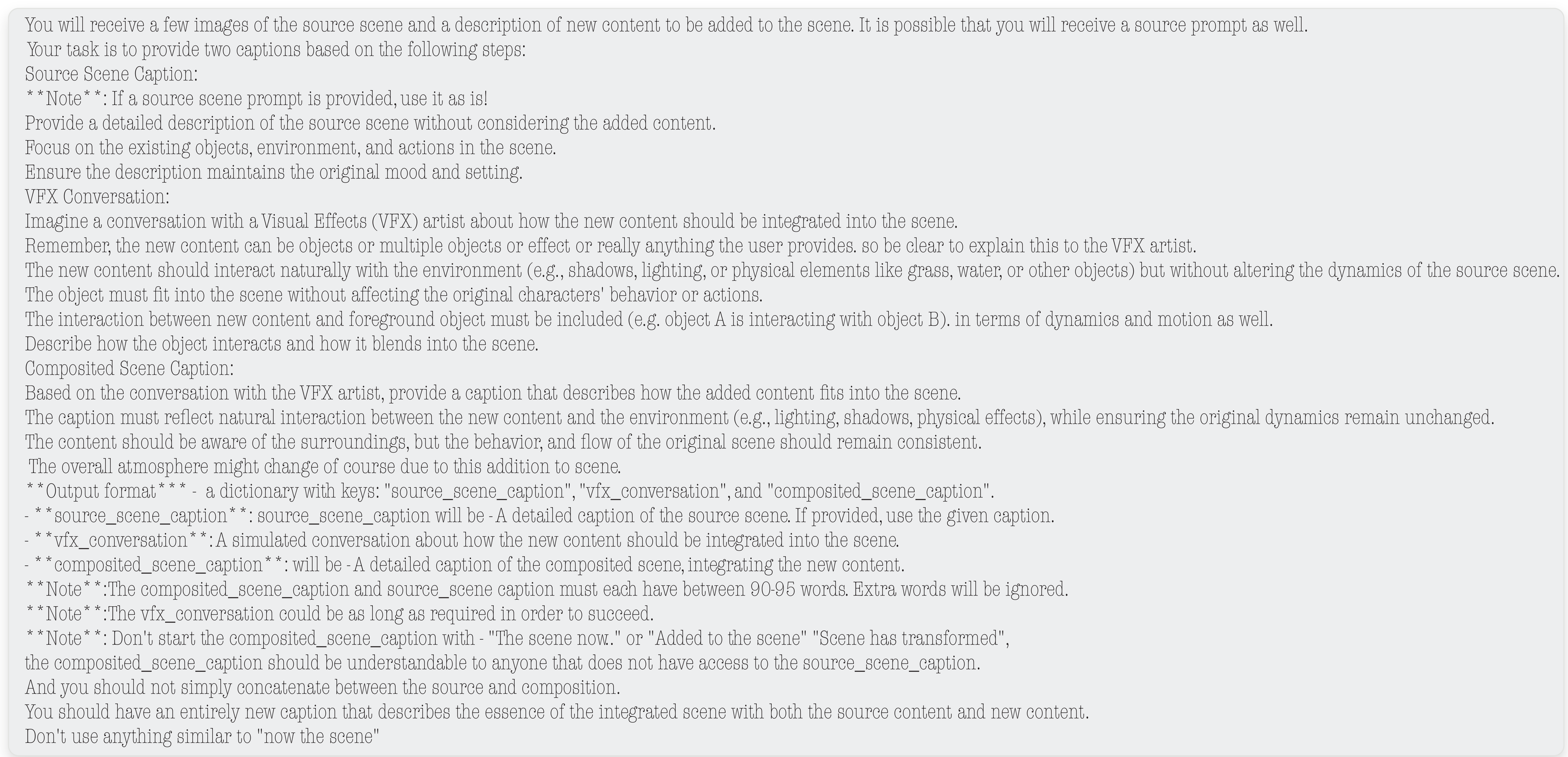}
    \caption{{\bf System Prompt We Input the Pre-trained VLM (VFX Assistant).} The system prompt contains guidelines to supply the VLM with the context of our task in order for it to interpret the user instruction, reason about the interaction, and accordingly, output a scene description that will be used for editing.}
    \label{fig:system_prompt}
\end{figure*}

Although the model is capable of providing descriptive captions of the source scene, in some cases, we observed that it fails to provide captions suitable for compositing VFX with the scene. To overcome this, we ask the model to imagine a conversation with a visual effects (VFX) artist to obtain a caption that would describe the composited scene correctly. In this conversation, GPT-4o will "consult" with a VFX artist about how the new content should be integrated into the scene. Based on their discussion, it will be asked to provide a caption that describes how the added content fits into the scene.
This results in text prompts that encourage the generated output video to include a natural interaction between the new content and the original environment. In this prompt, we also ask the VLM to provide a list of prominent foreground objects in the original video: $\mathcal{O}_{\text{orig}}$ and the object that will be added according to the edit prompt: $\mathcal{O}_{\text{edit}}$. The full prompt for the VLM is shown in Fig. ~\ref{fig:system_prompt}.

\begin{figure*}[htbp]
    \centering
\includegraphics[width=\textwidth]{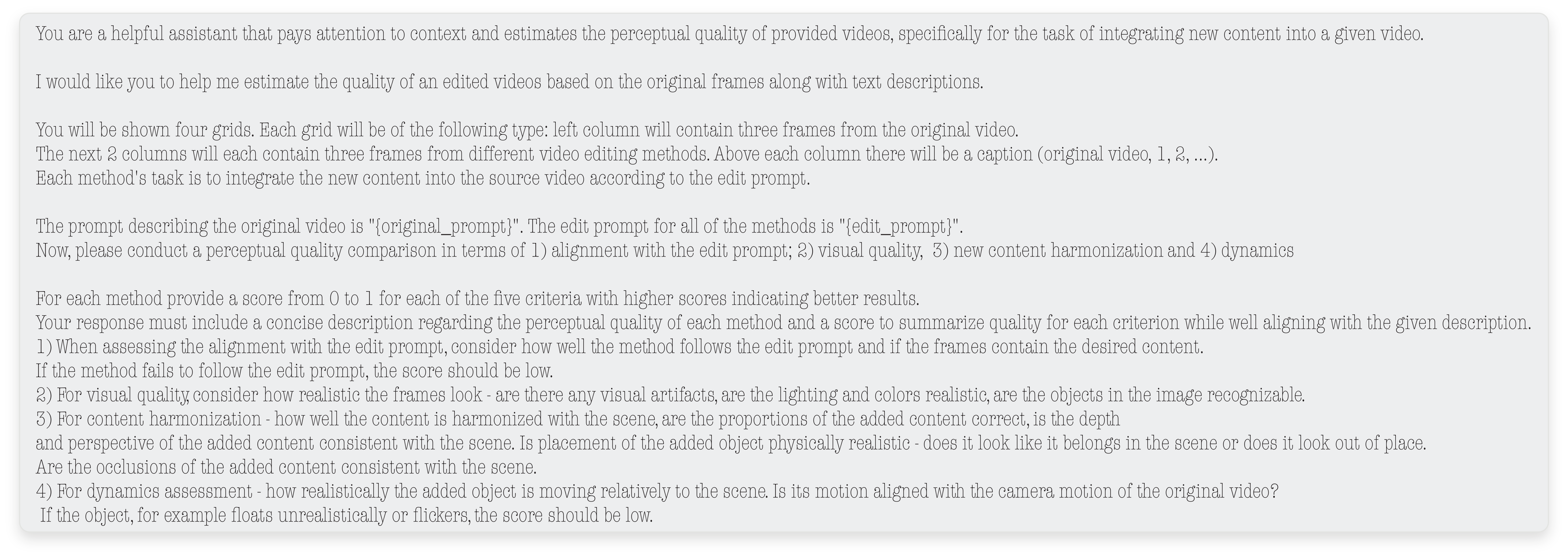}
    \caption{{\bf System Prompt for VLM-Based Evaluation.} containing the guidelines for the evaluation setup, where original and edited frames are compared against an edit prompt across four criteria: prompt alignment, visual quality, content harmonization, and dynamics.}
    \label{fig:vlm_evaluation}
\end{figure*}

In addition, as discussed in Sec. \ref{sec:quant}, we utilize the VLM for interpretable quality assessment. The full set of instructions for the VLM can be seen in Fig.~\ref{fig:vlm_evaluation}.

\subsection{Ablation of the VLM as a VFX Assistant Protocol}\label{app:ablate_vfx_protocol} 
\begin{figure}[htbp!]
\centering
\includegraphics[width=\linewidth]{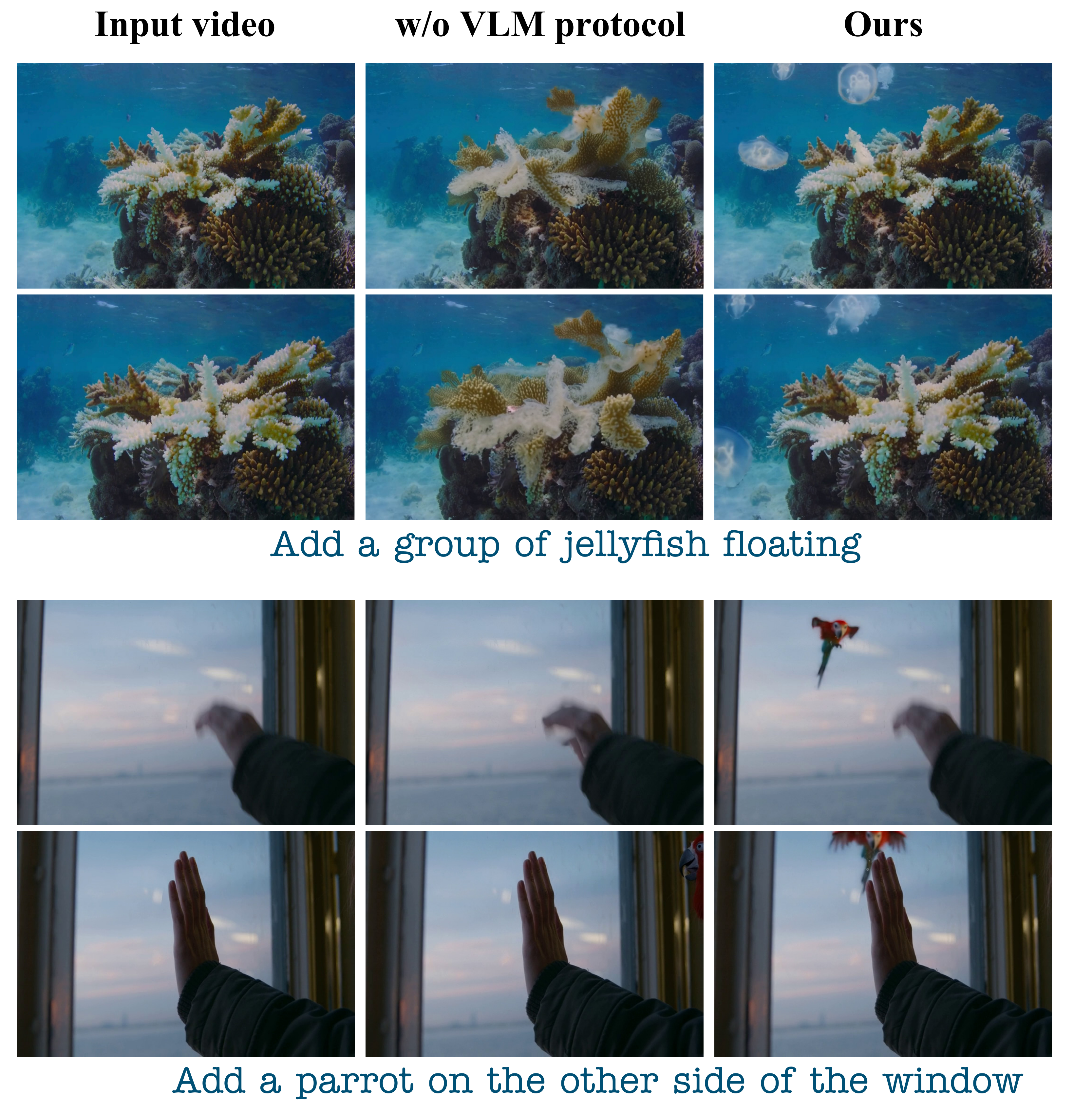}
        \vspace{-0.7cm}
\caption{{\bf Ablating VLM Protocol.} Without the protocol (middle), the VLM often yields an unsuitable prompt for adding the desired content, while our protocol (right), gives $\mathcal{P}_{\text{comp}}$ which allows for scene-aware additions to $\mathcal{O}_{\text{orig}}$ (see jellyfish and parrot behind the window). Additionally, as can be seen in Table \ref{tab:performance} in the main paper, Table \ref{tab:performance_3} and Fig. \ref{fig:metrics} from the SM, the metrics for this ablation are worse than those for our full method.}
\afterfigure
\label{fig:ablate_protocol}
\end{figure}

To validate the importance of our VLM protocol, we performed an ablation by prompting the VLM with a simplified system prompt, asking it to caption the original video and provide the edit prompt. As seen in Fig.~\ref {fig:ablate_protocol}, simplifying the system prompt results in misinterpretation of the instruction with respect to the scene or fails to add new content in general.

\section{VLM as Evaluation Metric Validation}\label{app:vlm_eval_metric} 
{\small
\begin{table}[ht!]
\caption{{\bf VLM Quality Evaluation with an alternative backbone (Claude Haiku 3.5).} We report text alignment, visual quality, edit harmonization, and dynamics score. The results closely match those obtained with GPT-4o in the main paper, showing consistent outcomes across all metrics and methods.}
\small
\setlength{\tabcolsep}{2.5pt}
\resizebox{\linewidth}{!}{%
\begin{tabular}{l|cccc}
\toprule
Method & \makecell{Text\\Align.} & \makecell{Visual\\Quality} & \makecell{Edit\\Harm.} & \makecell{Dynamics\\Score} \\
\midrule
Gen-3 & 0.517 & 0.499 & 0.412 & 0.445 \\
LORA finetuning & 0.807 & 0.755 & 0.722 & 0.719 \\
DDIM inv. sampling & 0.423 & 0.487 & 0.384 & 0.385 \\
SDEdit (0.9) & 0.769 & 0.734 & 0.697 & 0.682 \\
SDEdit (0.6) & 0.420 & 0.561 & 0.425 & 0.405 \\
FlowEdit & 0.670 & 0.651 & 0.591 & 0.605 \\
AnyV2V & 0.693 & 0.683 & 0.625 & 0.625 \\
I2VEdit & 0.769 & 0.747 & 0.697 & 0.698 \\
\midrule
w/o AnchorExtAttn & 0.656 & 0.620 & 0.547 & 0.556 \\
w/o Iterative Refinement & 0.704 & 0.672 & 0.631 & 0.634 \\
w/o VLM Protocol & 0.697 & 0.700 & 0.631 & 0.644 \\
Ours & \textbf{0.820} & \textbf{0.760} & \textbf{0.721} & \textbf{0.721} \\
\bottomrule
\end{tabular}}
\label{tab:performance_3}
\end{table}}

To show that our VLM-based evaluation aligns with human preferences, we performed the following assessments: 
Measured Pearson’s correlation coefficients between our VLM scores (Edit Harmonization and Dynamics) and the user study scores for Edit Harmonization, as both assess content integration quality. We observe a strong correlation $r=0.82$ ($p=0.025$) and $r=0.89$ ($p=0.007$). 
We compute Pearson’s correlation between the Text Alignment score and the CLIP Directional score -  a metric commonly used to assess text alignment and shown to correlate with human preference \cite{kim2025preservemodifycontextawareevaluation} - and found a strong correlation: r=0.81 (p=0.029). These strong correlations suggest that our metric aligns with human judgments in the context of our task.

We utilize the same VLM (GPT-4o \cite{achiam2023gpt4}) in our pipeline and evaluation. However, this does not introduce bias in our evaluation due to the separation between the internal representations of the T2V model and the VLM’s reasoning. During evaluation, the VLM does not “recognize” the video generated by the T2V model as its own, making the assessment unbiased and independent of its earlier role. We further validate this by evaluating the results with a different VLM (Claude Haiku 3.5 \cite{claudehaiku}) in Table \ref{tab:performance_3} and demonstrate that it provides similar scoring to that reported in the paper.

Our VLM-based evaluation metrics are inspired by recent work such as AutoVFX \cite{hsu2024autovfx} and \cite{wu2024comprehensivestudymultimodallarge}, which demonstrate the effectiveness of VLMs for automatic quality assessment tasks.


\begin{figure*}[t!]
    \centering
\includegraphics[width=\linewidth]{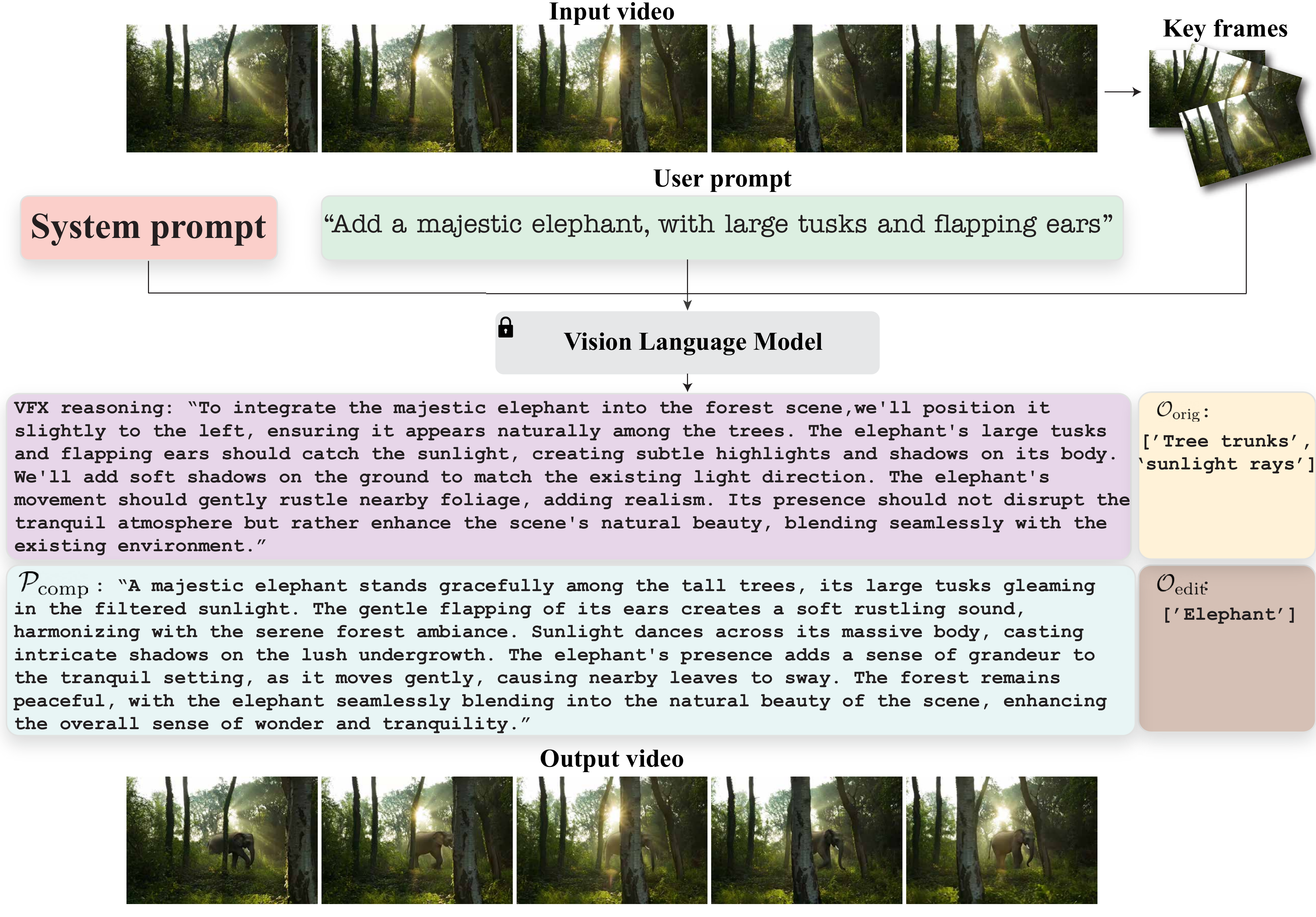}
     \vspace{-0.7cm}
    \caption{{\bf VLM as a VFX Assistant Overview.} Given instructions and an input video, the VLM (grey) interprets the interaction, reasons about the integration, and composes $\mathcal{P}_{\text{comp}}$ to guide the T2V generation. Inputs include: (1) a system prompt (red) guiding the VLM to produce VFX reasoning (purple), (2) a user prompt describing the content to integrate into the scene (green), and (3) key frames from the input video. The VLM produces VFX reasoning (purple) describing spatial placement, lighting, and motion for seamless integration, followed by $\mathcal{P}_{\text{comp}}$ (blue) captioning the target scene. Next, given the user prompt, the input video's key frames and $\mathcal{P}_{\text{comp}}$, the VLM identifies prominent elements from the original scene $\mathcal{O}_{\text{orig}}$ (yellow) and the new content to be added $\mathcal{O}_{\text{edit}}$ (brown). These outputs are used in our framework for the application of SAM, used to aid localization and harmonization of the new content. As can be seen in the output video, in this example, a majestic elephant is seamlessly integrated into the scene.}
    \label{fig:protocol}\vspace{-3mm}
\end{figure*}

\end{document}